\documentclass{article} 
\usepackage{iclr2025_conference,times}

\usepackage{amsmath,amsfonts,bm}

\def\eqref#1{equation~\ref{#1}}

\def\1{\bm{1}}

\DeclareMathAlphabet{\mathsfit}{\encodingdefault}{\sfdefault}{m}{sl}
\SetMathAlphabet{\mathsfit}{bold}{\encodingdefault}{\sfdefault}{bx}{n}

\usepackage{xspace}

\newcommand{\model}{{AskChart}\xspace}
\newcommand{\dataset}{{ChartBank}\xspace}

\usepackage{hyperref}
\usepackage{url}
\usepackage{booktabs}
\usepackage{multirow}
\usepackage{graphicx}
\usepackage{pifont}
\usepackage{ulem}
\usepackage{colortbl} 
\usepackage{caption}
\usepackage{wrapfig}
\usepackage{xcolor}  

\definecolor{lightblue}{rgb}{0.88,0.9,1}  
\usepackage{tcolorbox}
\newtcolorbox{egbox}[2][]{%
  colback=gray!10!white,
  colframe=gray!30!white,
  boxrule=0.2mm,
  left=0mm,
  right=0mm,
  top=0mm,
  bottom=0mm,
  coltitle=red!70!black,
  title={#2}, 
  #1
}

\renewcommand{\emph}[1]{\textit{#1}}

\title{AskChart: Universal Chart Understanding through Textual Enhancement}

\author{Xudong Yang$^1$,\ Yifan Wu$^{1\ast}$,\ Yizhang Zhu$^{1\ast}$, \ Nan Tang$^{1,2}$, \ Yuyu Luo$^{1,2\dagger}$ \\
\small{$^1$The Hong Kong University of Science and Technology (Guangzhou), Guangzhou, China} \\
\small{$^2$The Hong Kong University of Science and Technology, Hong Kong SAR, China} \\
}

\iclrfinalcopy 
\begin{document}

\maketitle

\renewcommand{\thefootnote}{\fnsymbol{footnote}}
    \footnotetext[1]{These authors contributed equally to this work.}
    \footnotetext[2]{Yuyu Luo is the corresponding author. E-mail: \texttt{yuyuluo@hkust-gz.edu.cn}}
\renewcommand{\thefootnote}{\arabic{footnote}}

\begin{abstract}
Chart understanding tasks such as ChartQA and Chart-to-Text involve automatically extracting and interpreting key information from charts, enabling users to query or convert visual data into structured formats. State-of-the-art approaches primarily focus on visual cues from chart images, failing to \textit{explicitly} incorporate rich textual information (e.g., data labels and axis labels) embedded within the charts. This textual information is vital for intuitive human comprehension and interpretation of charts. Moreover, existing models are often large and computationally intensive, limiting their practical applicability. 
In this paper, we introduce \model, a universal model that \textit{explicitly} integrates both \textit{textual} and \textit{visual} cues from charts using a Mixture of Experts (MoE) architecture. \model facilitates the learning of enhanced visual-textual representations of charts for effectively handling multiple chart understanding tasks, while maintaining a smaller model size. To capture the synergy between visual and textual modalities, we curate a large-scale dataset named \dataset with about 7.5M data samples, which helps align textual and visual information and facilitates the extraction of visual entities and text. To effectively train \model, we design a three-stage training strategy to align visual and textual modalities for learning robust visual-textual representations and optimizing the learning of the MoE layer. Extensive experiments across five datasets demonstrate the significant performance gains of \model in four chart understanding tasks. Remarkably, \model with 4.6B parameters outperforms state-of-the-art models with 13B parameters by \textbf{68.3\%} in Open-ended ChartQA and \textbf{49.2\%} in Chart-to-Text tasks, while achieving comparable performance in ChartQA and Chart-to-Table tasks.
\end{abstract}

\section{Introduction}

Charts are essential tools for data visualization, playing a crucial role in conveying complex data patterns in everyday applications~\citep{DBLP:journals/tkde/LuoQCTLL22, DBLP:journals/pacmmod/LuoZ00CS23}. Chart understanding tasks, including chart question answering (ChartQA)~\citep{chart_question_answering}, Chart-to-Text~\citep{kantharaj2022charttotext}, and Chart-to-Table translation~\citep{deplot}, aim to automate the interpretation and extraction of key information from charts, allowing users to query or convert visual data into structured formats.

With the advancement of multimodal large language models (MLLMs), recent studies aim to automatically perform various chart understanding tasks (e.g., ChartQA and Chart-to-Text) by pretraining MLLMs on large-scale chart-related corpus~\citep{masry2023unichart,chartllama,chartassisstant}.
For example, ChartAst~\citep{chartassisstant} is trained on a large-scale instruction-following chart-related corpus based on Donut~\citep{Donut} and SPHINX~\citep{lin2023sphinx} models, and demonstrates strong performance in ChartQA, Chart-to-Text and Chart-to-Table tasks.

\begin{figure*}[t!]
    \centering
\includegraphics[width=1.0\linewidth]{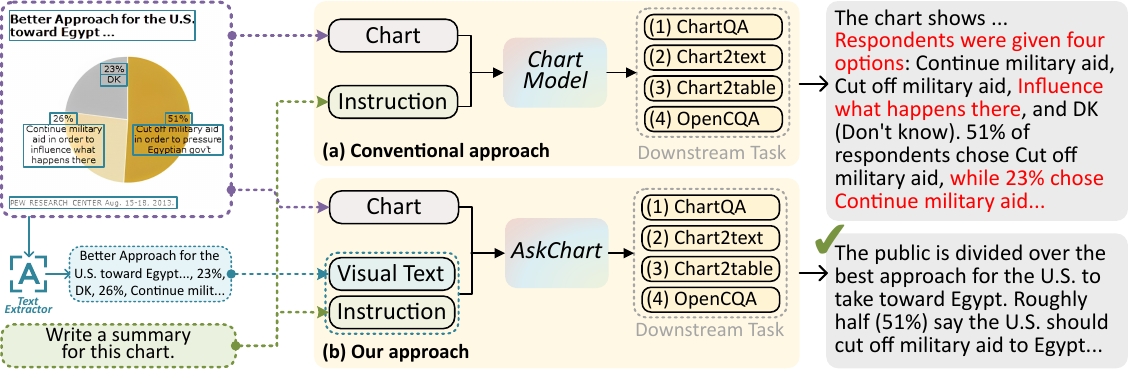}
    \vspace{-1.5em}
    \caption{Comparison between the conventional approach (specialized MLLMs) and our proposed method (\model) for chart understanding tasks. Our approach explicitly integrates both visual and textual information from charts, resulting in better performance in chart understanding tasks.}
    \label{movivation}
    \vspace{-1em}
\end{figure*}

Despite significant advancements, existing specialized MLLMs for chart understanding tasks predominantly rely on image-based representations, failing to \textit{explicitly} leverage the rich textual information embedded in charts~\citep{masry2023unichart,chartllama,chartassisstant}. This limitation reduces their effectiveness, particularly in tasks requiring precise interpretation of textual content. 
For example, as shown in Figure~\ref{movivation}(a), ChartAst~\citep{chartassisstant} misrepresents key facts, such as the percentage of slices of the pie chart, due to inadequate integration of textual data.

\paragraph{How do humans perform chart understanding?}
Humans naturally ``read'' and ``comprehend'' charts by integrating both \textit{textual} and \textit{visual} information~\citep{chartinsights, DBLP:journals/tvcg/SaketED19}.
When interpreting charts, people don’t focus solely on visual elements like bars or lines. Instead, they actively incorporate textual cues such as axes and data labels to form a complete understanding of the data being presented. These textual elements provide essential context, clarifying relationships between variables, and resolving ambiguities in the graphical representation~\citep{huang2024pixelsinsightssurveyautomatic}.

Inspired by this cognitive process,
our \textbf{key idea} is to \textit{explicitly} integrate textual information in chart understanding tasks, mimicking how humans interpret charts.
To achieve this, as shown in Figure~\ref{movivation}(b), our approach first employs a plug-in text extractor (e.g., OCR tools) to extract embedded textual information from the chart’s visual elements and then aligns both visual and textual modalities to learn more effective joint representations.
By explicitly combining both visual and textual cues, our approach could enable more accurate and comprehensive chart understanding, resulting in improved performance across various tasks such as Chart-to-Text and Open-ended ChartQA.

\paragraph{Challenges.} Directly employing OCR tools to extract text from charts often results in errors such as misrecognition, incomplete extraction, or misalignment, particularly when dealing with complex chart structures. This presents the first challenge: 
\textbf{\textit{(C1: Alignment Challenge)}} How to accurately align noisy OCR text with the corresponding visual components of the chart, enabling the model to learn meaningful joint representations and avoid misinterpretation?
\textbf{\textit{(C2: Architectural Challenge)}} How can we design a flexible and efficient architecture that can dynamically adapt to different chart types and tasks, effectively integrating visual and textual cues to optimize performance?
\textbf{\textit{(C3:~Dataset Challenge)}} Existing datasets lack comprehensive training data that integrates both structural visual elements and textual information for chart understanding tasks.

\paragraph{Our Methodology.} In response to these challenges, we introduce \model, a universal model that \textit{explicitly} integrates both \textit{textual} and \textit{visual} cues from charts using a sparse Mixture of Experts (MoE) architecture to tackle multiple chart understanding tasks effectively.
Specifically, \model utilizes a plug-in text extractor to extract textual information from charts, which is processed alongside user instructions via text encoders. In parallel, the visual encoder captures structural and visual chart information. The attention mechanism in LLMs integrates these components, while visual-textual alignment learning ensures the noisy extracted text is accurately aligned with its corresponding visual elements (addressing~\textit{\textbf{C1}}).
To effectively handle diverse chart types and tasks without compromising on performance and efficiency, 
\model employs MoE layers, which allows for sparse computation, activating only the relevant experts and reducing unnecessary overhead by dynamically distributing tasks among specialized experts (addressing~\textit{\textbf{C2}}).

To address the third challenge (\textit{\textbf{C3}}), we construct \dataset, a large-scale dataset consisting of approximately 7.5 million samples that integrates both visual and textual elements from various chart-related tasks.
\dataset consists of three specialized datasets:
\textit{(a) the OCR-aware Data Prompt Dataset}: Aligns textual and visual information by featuring both single-turn and multi-turn instruction-following tasks, such as OpenCQA, Chart-to-Table, and chart summarization.
\textit{(b) Visual Prompt Dataset}: Comprising three types of chart question-answering tasks, i.e., reasoning, search, and data retrieval, where answers are visually highlighted using various prompt types (e.g., ellipses, bounding boxes, triangles) to enhance feature learning on chart images.
\textit{(c) the Chart-to-Table Instruction-Following Dataset}: Facilitates table and text extraction from charts.

\paragraph{Contribution.} Our contributions can be summarized as follows:

(1) {\bf New Methodology.} 
We propose \model, a lightweight 
model that explicitly integrates both textual and visual cues through MoE layers. We employ a three-stage training strategy with tailored pretraining objectives to enhance its performance across diverse chart understanding tasks.

(2) {\bf New Dataset.}
We introduce \dataset, a large-scale dataset with approximately 7.5 million samples, comprising three specialized sub-datasets: the Visual Prompt Dataset, the OCR-aware Instruction-Following Dataset, and the Chart-to-Table Instruction-Following Dataset.
	
(3) {\bf Extensive Experiments.}
Our approach achieves new state-of-the-art performance across multiple benchmarks. \model outperforms larger models, such as those with 13B parameters, by 68.3\% in Open-ended ChartQA and 49.2\% in Chart-to-Text tasks, while delivering comparable results in ChartQA and Chart-to-Table tasks. We make both code and datasets publicly available at {\url{https://github.com/Sootung/AskChart}}.

\section{Related Work}

{\bf Chart Understanding.}
In chart understanding, key tasks have emerged, each focusing on interpreting and reasoning over chart data~\cite{DBLP:journals/tvcg/LuoTLTCQ22, DBLP:conf/sigmod/Luo00CLQ21, DBLP:journals/vldb/QinLTL20}. ChartQA~\citep{chart_question_answering, xu2023chartbench} involves answering questions related to both the content and structure of charts, requiring models to extract insights from graphical elements. The Chart-to-Table~\citep{deplot} task converts visual chart data into structured tables for easier analysis, while Chart-to-Text~\citep{kantharaj2022charttotext} generates descriptive text from chart information. Complex tasks like Open-ended ChartQA (Open~CQA)~\citep{kantharaj2022opencqa} demand higher-level reasoning beyond fact retrieval. Our \model is designed to handle these four core chart understanding tasks.

{\bf MLLMs for Chart Understanding.}
MLLMs like LLaVA~\citep{llava} and BLIP2~\citep{li2023blip2} have excelled in chart understanding tasks by leveraging abundant natural image datasets~\citep{Conceptual,coco,llava}. 
However, high-quality pre-training datasets for charts are still underexplored. Existing methods like UniChart~\citep{masry2023unichart} expand task types but struggle with complex reasoning. Models like ChartLLaMA~\citep{chartllama}, ChartAssistant~\citep{chartassisstant}, ChartGemma~\citep{masry2024chartgemmavisualinstructiontuningchart}, and ChartInstruct~\citep{masry2024chartinstructinstructiontuningchart} aim to address chart reasoning and editing tasks, while ChartMoE~\citep{xu2024chartmoemixtureexpertconnector} improves multimodal input handling. However, open-ended tasks like OpenCQA~\citep{kantharaj2022opencqa} remain challenging. We propose \model with a visual-textual alignment pre-training approach that achieves state-of-the-art results in OpenCQA by better aligning visual chart structure with textual information of charts.

\begin{figure*}[t] 
	\vspace{-1em}
	\centering
\includegraphics[width=1.0\linewidth]{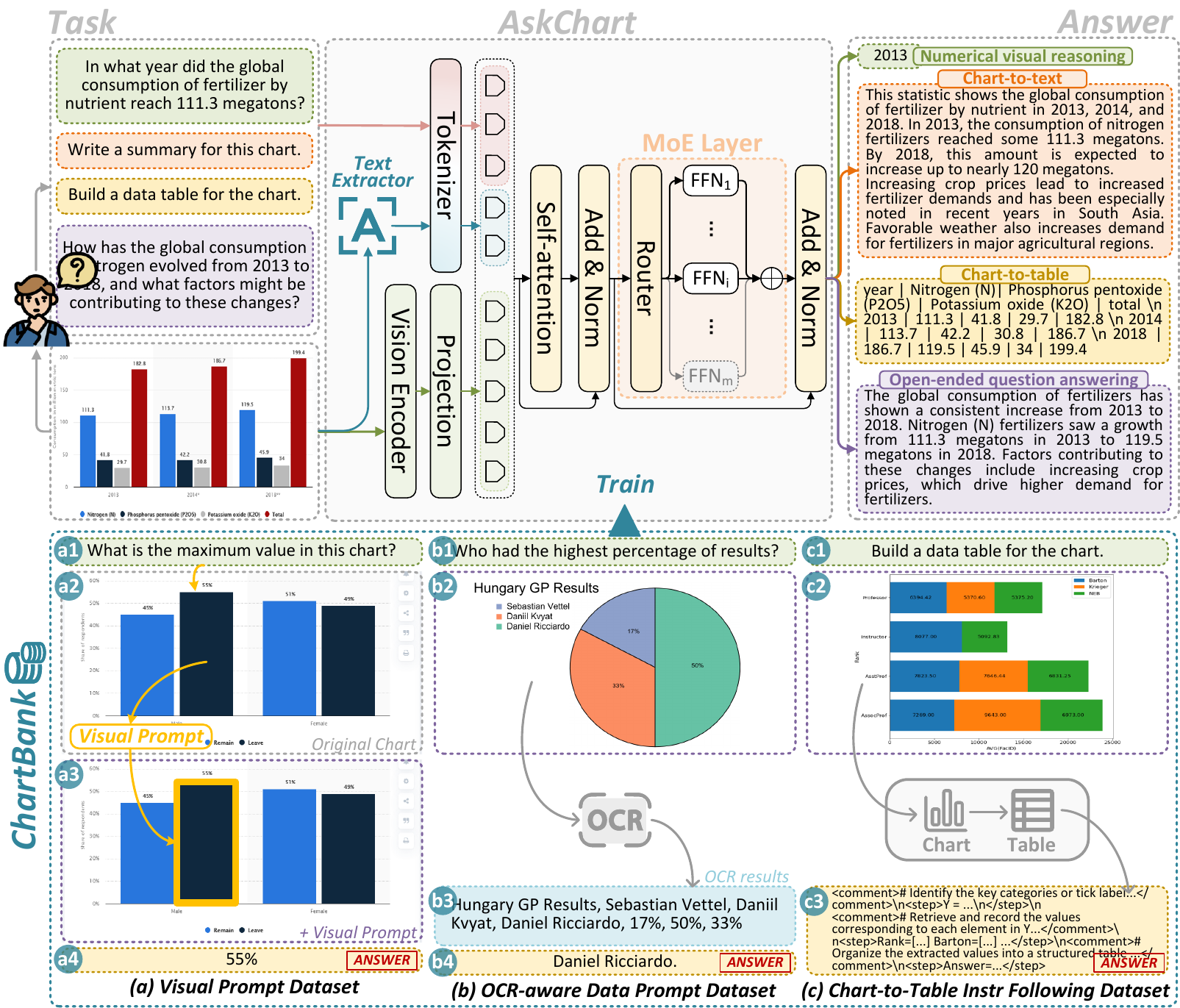}
	\vspace{-1.25em}
	\caption{The framework of \model. The upper part shows the processing pipeline and \model structure while the lower part shows examples in \dataset for pretraining. We newly curate three datasets: (a) Visual Prompt Dataset, (b) OCR-aware Data Prompt Dataset, and (c) Chart-to-Table Instruction Following Dataset. For \dataset examples in lower part, blocks in \textcolor[RGB]{0,128,0}{green} indicate \textit{tasks} (a1, b1, c1); blocks with \textcolor[RGB]{126,100,158}{purple borders} indicate \textit{input charts} (a3, b2, c2); block in \textcolor[RGB]{146,205,220}{blue} is the \textit{OCR result} (b3); blocks in \textcolor[RGB]{255,192,0}{yellow} indicate \textit{answers} (a4, b4, c3).}
	\label{framework}
	\vspace{-.3cm}
\end{figure*}

{\bf Visual-Textual Alignment Learning.}
Recent MLLMs~\citep{llavar,moe-llava,chartllama} like LLaVA~\citep{llava} use single-turn conversations between humans and an assistant to briefly describe natural images. However, for charts, descriptions often include content that visual entities alone cannot capture (e.g., the semantic context of the chart)~\citep{kantharaj2022charttotext}, which results in relatively noisy data for alignment tasks. Models like PresSTU~\citep{PreSTU}, PaLI~\citep{pali}, and LLaVAR~\citep{llavar} utilize noisy OCR-generated text as ground-truth prediction answers to enhance the model's text comprehension capabilities. Nevertheless, this noisy data remains insufficient for achieving robust alignment~\citep{xu2020layoutlm, ren2016faster_R-CNN}. LayoutLM~\citep{xu2020layoutlm} relies on object detection networks~\citep{ren2016faster_R-CNN}, which tend to underperform in charts that are rich in structural visual units, as they struggle to compute the patch-OCR loss to align vision and text. Similarly, ChartBERT~\citep{xu2023chartbench}, though using OCR-generated text, lacks the ability to effectively represent image and text information jointly. Limited approaches incorporate visual text as input for visual instruction fine-tuning.
Our fundamental premise is to explicitly integrate visual-textual information with the user instruction, and then process them in parallel with the chart tokens through the training process of our \model.

\section{\model Model}
\label{sec:model}

We will first present the architecture of \model (Section~\ref{sub:arct}). We will then introduce the training objectives (Section~\ref{sub:training_obj}) and finally elaborate on the training strategy (Section~\ref{sub:training_strategy}).

\subsection{AskChart Architecture}
\label{sub:arct}

{\bf Overall Architecture.}
As shown in Figure~\ref{framework}, the architecture of \model is designed to efficiently integrate both textual and visual information from charts. \model incorporates a text extraction module $\varphi(\cdot)$, which retrieves textual data from charts, alongside user instructions processed through a word embedding layer $g_t(\cdot)$. Simultaneously, a vision encoder $g_v(\cdot)$, captures the structural and visual elements. The extracted multimodal features are then aligned using a projection layer $proj(\cdot)$, and passed to an LLM, $f_\theta(\cdot)$. The LLM is enhanced with the MoE architecture, which dynamically allocates specialized experts to specific tokens. This design not only ensures efficiency and scalability but also enables the model to effectively manage the complex interactions between visual and textual modalities, all while maintaining a lightweight computational footprint.

To achieve a lightweight model, we adopt a tiny LLM (e.g., Phi) as a replacement for larger models like Vicuna~\citep{vicuna} and LLaMA~\citep{touvron2023llama}. Both the image encoder and LLM are built upon one of the recent state-of-the-art lightweight MLLMs, MoE-LLaVA~\citep{moe-llava}.
Given an input chart $\mathbf{X}_v$, the vision encoder processes the chart and generates a sequence of visual tokens. These tokens are then passed through a projection layer, which maps the visual tokens into language embedding tokens $\mathbf{H}_v$. Simultaneously, the text extractor processes the chart to extract visual text from the image, which is then combined with the user's instruction. Both the visual text $\mathbf{X}_o=\varphi(\mathbf{X}_v)$ and instructions $\mathbf{X}_t$ are passed through $g_t(\cdot)$ to generate visual-text sequence tokens $\mathbf{H}_o$ and instruction sequence tokens $\mathbf{H}_t$. Since the visual text is essentially textual information, we utilize the same text encoder for this task to simplify the process. Consequently, the token sequences $\mathbf{H}_v$, $\mathbf{H}_t$, and $\mathbf{H}_o$ are concatenated and fed into the LLM, which uses MoE layers to replace the traditional feed-forward networks (FFNs). Each MoE block consists of a learnable router and multiple FFNs. The entire model workflow can be formally defined by the following equations:
\begin{equation}
    \mathbf{H}_v = proj(g_v(\mathbf{X}_v)); \mathbf{H}_t = g_t(\mathbf{X}_t); \mathbf{H}_o = g_t(\mathbf{X}_o),
\end{equation}
\begin{equation}
    \mathcal Y = f_\theta([\mathbf{H}_v;\mathbf{H}_t;\mathbf{H}_o]),
\end{equation}
where $\mathcal Y$ is the output answer. 

{\bf Text Extractor.}
The text extractor is designed to accurately recognize task-agnostic visual text in charts with varying resolutions. Although some OCR-free vision encoders~\citep{Donut,xu2020layoutlm} trained on domain-specific data excel at understanding scene text, their generalization ability is limited, particularly when dealing with visual text in charts that vary in font size and style. Additionally, compared to some open-source OCR tools, these models often have a much larger number of parameters, making them difficult to deploy and fine-tune in resource-constrained environments. Therefore, we adopt a lightweight OCR tool, PaddleOCR~\citep{PaddleOCR}, as the text extractor. Given a chart, PaddleOCR sequentially extracts the text by scanning from the top-left corner to the bottom-right corner of the image. The recognized visual text $\mathbf{X}_o$, which forms part of the LLM prompts used during both training and inference, is then concatenated with the user instruction $\mathbf{X}_t$.

\subsection{Training Objectives}
\label{sub:training_obj}

We perform instruction-tuning of \model. Specifically, we train the LLM with MoE and the Vision Encoder in \model on the prediction tokens, using both the original~\citep{moe-llava} auto-regressive loss $\mathcal{L}_{reg}$ and an auxiliary loss $\mathcal{L}_{aux}$~\citep{Switch-Transformers} which encourages the router to efficiently balance the load across multiple experts. The combined objective can be expressed as:
\begin{equation}
	\mathcal{L} = \mathcal{L}_{reg} + \lambda\mathcal{L}_{aux} ,
\end{equation}
where $\lambda$ is a balancing factor that controls the contribution of the auxiliary loss $\mathcal{L}_{aux}$. 

Given a sequence of length $L$, the auto-regressive loss of the target answers $\mathcal{Y}_{a}$ is defined as,
\begin{equation}
	\mathcal{L}_{reg} = - \sum_{i=1}^{L} \log p_{\theta} \left( y_i \mid \mathbf{X}_v, \mathbf{X}_o, \mathbf{X}_{t,<i}, \mathcal{Y}_{a,<i} \right)~,
\end{equation}

where $\theta$ is the trainable parameters, $y_i$ is the current prediction token. 

For $N$ experts, the auxiliary loss $\mathcal{L}_{aux}$ is computed as,
\begin{equation}
	\mathcal{L}_{aux} = N \cdot \sum_{i=1}^{N} \mathcal{F}_{i} \cdot \mathcal{P}_{i}~,
\end{equation}
where $\mathcal{F}$ is the fraction of tokens processed by expert $i$, and $\mathcal{P}$ represents the portion of the router probability assigned to expert $i$, which can be defined as:
\begin{equation}
	\mathcal{F}_{i} = \frac{1}{L} \sum_{i=1}^{L} \mathbf{1}\left\{\arg\max p(x) = i \right\};~~
	\mathcal{P}_{i} = \frac{1}{L} \sum_{i=1}^{L} p_i(x)~.
\end{equation}

\vspace{-1em}
\subsection{Training Strategy}
\label{sub:training_strategy}

To effectively train \model, we adopt a three-stage training strategy designed to align visual and textual modalities in charts, ensuring the model learns robust visual-textual representations. This strategy also fine-tunes the MoE layers to handle diverse chart understanding tasks efficiently. 
Throughout these stages, we employ multi-task tuning based on the \dataset dataset (will be introduced in Section~\ref{sec:dataset}). Unlike existing MLLMs~\citep{llava,moe-llava,chartassisstant}, which typically freeze the vision encoder during training, we find that unfreezing the vision encoder across all stages significantly improves performance in chart understanding tasks.

Table~\ref{tab:pretrain_dataset} in the Appendix shows the tasks and datasets used across the different training stages.

{\bf Stage I: Visual-Textual Alignment.} 
Effective chart understanding requires the model to establish a clear relationship between the chart’s visual representation and its corresponding textual information. The goal of this stage is to accurately align noisy OCR-extracted text with the visual elements of the chart.
To achieve this, we use Chart-to-Table translation as a pretraining task, similar to approaches used in ChartAst~\citep{chartassisstant} and Matcha~\citep{liu2022matcha}. The vision encoder and projection layer are trained to map image tokens into pseudo-text tokens. During this phase, we utilize relatively noisy chart-table pairs, where some of the underlying data tables are estimated based on the graphical marks (e.g., bars) as a percentage of the chart's plot area~\citep{masry2023unichart}. Although this introduces some noise, we mitigate it with high-quality datasets during fine-tuning, effectively aiding the model in aligning charts with their corresponding tables.

{\bf Stage II: Multi-task Instruction Tuning.} 
This stage aims to enable the model to generalize across various chart understanding tasks and diverse user instructions.
As shown in Table~\ref{tab:pretrain_dataset}, a key task is chart summarization, where the model generates summaries of chart content based on different user instructions, enhancing its ability to produce varying levels of detail.
Specifically, Numerical and visual reasoning tasks go beyond the template-based reasoning seen in UniChart~\citep{masry2023unichart}, by incorporating multi-turn conversations, covering sub-tasks like chart structural understanding, data retrieval, and mathematical reasoning. 
The open-ended ChartQA~\citep{kantharaj2022opencqa} task involves high-level questions requiring reasoning and explanatory answers. To address these, the model must comprehend visual text, demanding both perceptual and cognitive understanding. In contrast, low-level ChartQA tasks focus on specific goals such as reasoning, searching, and data retrieval. Each chart is marked with visual prompts to guide the model toward specific, highlighted areas of the image, improving task focus and accuracy.

{\bf Stage III: Fine-tuning with Mixture of Experts.}
To mitigate the learning difficulty associated with the sparse model architecture, we initialize the weights in the third stage using those from the second stage. When tokens are fed into the MoE layers, the router activates the top-$k$ experts to handle the tokens, and their outputs are combined using a weighted sum based on the router's weights. This mechanism helps distribute the computational load across multiple experts, improving the model's efficiency. In this stage, we fine-tune the model on tasks that are highly relevant to downstream tasks. Recognizing the challenges of translating charts to tables, we introduce a Chain-of-Thought (CoT)-based~\citep{wei2022Chain-of-thought} translation task. This task requires the model to generate a step-by-step reasoning process (CoT) rather than producing a direct answer. By generating CoT answers, the model is encouraged to explicitly demonstrate its reasoning pathway, which leads to more accurate and interpretable results, particularly for complex Chart-to-Table translation tasks.

\vspace{-.5em}
\section{\dataset Dataset}
\label{sec:dataset}
\vspace{-.5em}

To enhance \model's chart understanding capabilities, we curate \dataset, comprising three specialized datasets alongside existing work: (1) the Visual Prompt Dataset, (2) the OCR-aware Data Prompt Dataset, and (3) the Chart-to-Table Instruction-Following Dataset. 

{\bf \dataset Overview.}
Figure~\ref{framework} illustrates examples from our  \dataset, and Appendix~\ref{sta_chartbank} provides a summary of the \dataset statistics.
Specifically, the Visual Prompt Dataset and OCR-aware Data Prompt Dataset cover 6 representative chart types: pie, common bar, stacked bar, grouped bar, common line, and grouped line charts. Among these types of charts, the common bar and common line both have only one category of data, while the stacked bar, grouped bar, and grouped line all have multiple categories of data. The Chart-to-Table Instruction Following Dataset additionally involves scatter plots. We transform all datasets, including datasets introduced by us and training sets of existing UniChart~\citep{masry2023unichart}, ChatQA~\citep{masry2022chartqa}, OpenCQA~\citep{kantharaj2022opencqa}, Chat-to-text~\citep{kantharaj2022charttotext} datasets, into an instruction-following format for pretraining. As shown in Appendix~\ref{appsub:instruction}, we design various instruction templates for random selection to increase language diversity. All the instruction-following datasets are used during the pretraining stages as illustrated in Table~\ref{tab:pretrain_dataset}. 
Next, we will introduce the design consideration construction pipelines for each specialized dataset in \dataset. For more details, please refer to Appendix~\ref{app:dataset}.

\subsection{Visual Prompt Dataset}
\vspace{-.5em}

Region understanding capabilities are crucial in chart understanding, as questions often target only particular elements, like individual bars in a bar chart. We also aim to strengthen the MLLM's numerical visual reasoning to understand relationships among numerical values. Therefore, we develop and incorporate the Visual Prompt Dataset for second-stage pretraining, as shown in Figure~\ref{framework}(a).

{\bf Construction.} 
Charts in ChatQA \citep{masry2022chartqa} are utilized as the foundation to construct the Visual Prompt Dataset. Firstly, we carefully design question templates (Appendix Table~\ref{tab:question_temp_visual}) to be used in question generation for four tasks: (1) reasoning, (2) extremum, (3) determining range, and (4) data retrieval. 
Subsequently, for each chart, we randomly select elements to generate questions and record their bounding box indices, thereby overlapping the visual prompt using ViP-LLaVA \citep{cai2024vipllava}. Charts unable to be visually prompted accurately by ViP-LLaVA, like involving correlation and distribution tasks, will be deemed unsuitable and consequently excluded. For diversity, we randomly select three types of visual prompts from a set of four (namely arrow, ellipsis, bounding box, and triangle) for each question, yielding 417,780 (Chart, Question, Answer) pairs ultimately. Figure~\ref{framework}-a2, a3 illustrates an example with the rectangle visual prompt.

\subsection{OCR-aware Data Prompt Dataset}

As mentioned, the weakness in text capture and utilization is a bottleneck limiting MLLMs' chart understanding capabilities. We aim to enhance MLLMs' such capabilities by providing richer and denser textual information aligned with the features in charts. Also, multi-turn question-answering examples are included to enable the model to better fit real-world scenarios. Therefore, we introduce the OCR-aware Data Prompt Dataset in the second-stage pretraining, as shown in Figure~\ref{framework}(b).

{\bf Construction.} 
The OCR-aware Data Prompt Dataset includes two parts: single-turn and multi-turn instruction-following data, with each example comprising four essential elements: questions (Figure~\ref{framework}-b1), charts (Figure~\ref{framework}-b2), OCR results (Figure~\ref{framework}-b3), and answers (Figure~\ref{framework}-b4). For both single-turn and multi-turn examples, we employ PaddleOCR to extract textual information from the input charts to obtain OCR results. The single-turn instruction-following data is directly derived from UniChart ~\citep{masry2023unichart} through format transformation, containing 6,791,230 examples.
For multi-turn data, we utilize charts in UniChart accompanied by original tables, serving as the foundation for generation. First, we prompt ChatGPT~\citep{ouyang2022training} to identify and summarize the common question types in PlotQA~\citep{methani2020plotqa} templates, which encompass three question-answering task categories: structural understanding, data retrieval, and mathematical reasoning. To enhance the effectiveness and accuracy of question and answer generation, we provide ChatGPT with sequenced original tables instead of charts. Then ChatGPT is prompted to synthetically generate two to three rounds of questions and answers, guided by identified question types (prompts in Appendix Table~\ref{tab:prompt_multi_turn}). Finally, we obtain 189,747 multi-turn examples.

\subsection{Chart-to-Table Instruction Following Dataset}

To improve \model's ability to comprehensively extract and understand information from charts, we propose CoT based the Chart-to-Table Instruction Following Dataset for the third-stage fine-tuning, as shown by the example in Figure~\ref{framework}(c).

{\bf Construction.} 
We construct a large amount of high-quality (chart, COT annotated table) pairs by converting tables into charts with CoT ground-truth answers (see Appendix~\ref{cot_dataset}).
To this end, we first utilize widely used Text-to-SQL datasets,
Spider~\citep{spider} and BIRD~\citep{bird}, which contain 1,020 and 1,460 tables on 138 and 37 domains, respectively, as the base table. we first employ the automatic visualization system, DeepEye~\citep{luo2018deepeye}, to  recommend good charts for these tables. Subsequently, we use Matplotlib to render the charts. Finally, we have a total of 61,472 (chart, table) pairs for forming our Chart-to-Table Dataset.

\section{Experiments}

\subsection{Experimental Setup}

{\bf Datasets and Tasks.}
We evaluate \model against state-of-the-art (SOTA) methods on four chart understanding tasks using various widely-used benchmarks. For  ChartQA, we use the ChartQA benchmark~\citep{masry2022chartqa}, which focuses on visual and logical reasoning, where each question typically has a single word or numerical answer. This benchmark also includes the Chart-to-Table translation task, for which we follow the evaluation methodology from prior work. Additionally, we assess the model’s performance in the chart summarization task using the Chart-to-Text benchmark~\citep{kantharaj2022charttotext}. For Open-ended ChartQA, we evaluate using the OpenCQA benchmark~\citep{kantharaj2022opencqa}, where questions require more explanatory and detailed answers.

{\bf Evaluation Metrics.}
We adopt evaluation metrics from prior studies~\citep{masry2022chartqa}. For ChartQA, we use relaxed accuracy (RA), allowing a 5\% margin of error for numerical answers and exact matches for textual answers. For Chart-to-Table, we report RMS-F1 scores based on the DePlot framework~\citep{deplot}. Both the Chart-to-Text task and OpenCQA are evaluated using BLEU scores~\citep{2018bleu}, consistent with previous works~\citep{masry2023unichart,liu2022matcha}.

{\bf Baselines.}
We compare \model with three types of baselines.
\textbf{(i)~General-Purpose MLLMs.} These models are designed for diverse multimodal tasks and have demonstrated strong image understanding capabilities. We included {Blip2}~\citep{li2023blip2}, {SPHINX}~\citep{lin2023sphinx}, and {Qwen-VL}~\citep{bai2023qwen} as representative models to assess their applicability to chart-specific tasks.  
\textbf{(ii)~Specialist Chart Models.} This category focuses on models tailored for document and chart comprehension. In addition to {DePlot}~\citep{deplot} and {OneChart}~\citep{chen2024onechart}, We considered {UniChart}~\citep{masry2023unichart} and {MatCha}~\citep{liu2022matcha}, which build on the architectures of {Donut}~\citep{Donut} and {Pix2Struct}~\citep{pix2struct} respectively, known for their strengths in document understanding. Additionally, we evaluated {Chart-T5}~\citep{chart-t5}, an enhanced version of the text-centric T5 model~\citep{T5}, which has been adapted for solving chart-related language tasks.  
\textbf{(iii)~Chart MLLMs.} These models are explicitly designed for chart-related tasks and leverage popular vision-language model architectures to achieve state-of-the-art performance. We selected {ChartInstruct}~\citep{masry2024chartinstruct}, {ChartLLaMa}~\citep{chartllama}{TinyChart}~\citep{zhang2024tinychart} and {ChartAst}~\citep{chartassisstant}, which demonstrate advanced capabilities in various chart comprehension challenges.

{\bf Implementation Details.} \model integrating SigLIP~\citep{siglip} as the vision encoder and Phi-2~\citep{phi} as the language model. We trained all models using 8 A100 GPUs. 
Table~\ref{tab:pretrain_dataset} shows all datasets used for training. For Stage~I, we trained the model for 1 epoch with a learning rate of 1e-3 and a batch size of 32 per GPU. For Stage II and Stage III, we fine-tuned the model for 1 and 6 epochs, respectively, with a learning rate of 2e-5 and a batch size of 16 per GPU. Please refer to Appendix~\ref{app:training} for more details.

\subsection{Main Results}

\begin{figure*}[t!]
    \vspace{-1em}
    \hspace*{-1em}
    \includegraphics[width=1.02\linewidth]{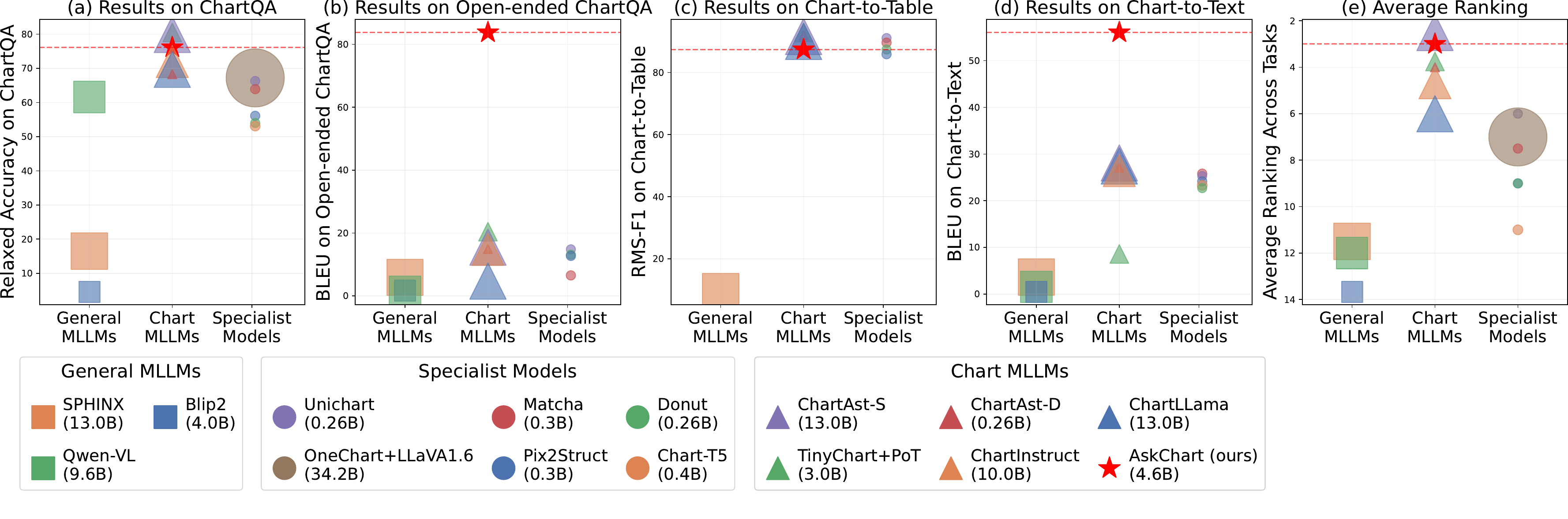}
    	\vspace{-2em}
    \caption{Evaluation results on chart-related benchmarks. The size of the markers represents the size of the corresponding models, with larger markers indicating larger model sizes. Our \model is represented as a red star, demonstrating its good performance across various tasks.
    }
    \label{fig:main result}
\end{figure*}


Figure~\ref{fig:main result} shows a comparison of AskChart with SOTA models across four chart understanding benchmarks. Remarkably, AskChart outperforms the current state-of-the-art methods by 68.3\% and 49.2\% (on the Pew sub-dataset), and 6.7\% (on the Statista sub-dataset) in the open-ended ChartQA and chart-to-text tasks, respectively. This demonstrates that the lightweight AskChart (4.6B parameters) achieves competitive results on ChartQA and Chart-to-Table tasks, comparable to the performance of ChartAst-S (13B parameters)~\citep{chartllama}.
We observe that existing models struggle to effectively handle long text generation tasks, such as open-ended ChartQA, which requires generating explanatory answers by reasoning with chart content, and Chart-to-Text, which demands an integrated understanding of visual and textual information in charts. Due to its explicit text enhancement and multitask training, AskChart performs joint visual and explicit text representation, and its MoE architecture enables a single token to be processed by different experts, with weighted outputs providing a more robust representation. This capability allows it to effectively address such complex tasks.
Moreover, AskChart demonstrates significant advantages in tasks that demand both text recognition and generation. Unlike certain models, such as UniChart~\citep{masry2023unichart} and MatCha~\citep{liu2022matcha}, which require fine-tuning for each downstream task to achieve optimal performance and often rely on separate models for different tasks, AskChart serves as a universal solution capable of addressing diverse requirements without task-specific fine-tuning.

Additionally, we conducted an error analysis based on chart types and question types (see Appendix~\ref{error-analysis}). From the accuracy distribution across different chart types, it is evident that the performance of AskChart is almost unaffected by the chart type, with comparable performance across various chart categories. To further analyze performance from the perspective of question types, we randomly selected 1,108 human-written questions. The model's performance was notably lower on data retrieval and compositional tasks that require multi-step reasoning, indicating that the vision encoder struggles with understanding chart values, while the large language model exhibits limitations in mathematical reasoning. These challenges primarily stem from the model's susceptibility to hallucinations in fine visual elements and its insufficient capacity for numerical representation.

\subsection{Further Study}

\begin{table}[t!]
\centering
\caption{Evaluation results on ChartInsights benchmark.}
\vspace{-.6em}
	\label{tab:askchart_on_chartinsights}
	\fontsize{8pt}{10pt}\selectfont
    \resizebox{\linewidth}{!}
{
	\begin{tabular}{lcccccccccccc}
		\toprule
		\multirow{2}{*}{Model}& \multirow{2}{*}{Size} & \multicolumn{4}{c}{Analysis} & \multicolumn{3}{c}{Search} & \multicolumn{3}{c}{Query} & \multirow{2}{*}{Overall ($\%$)} \\
		\cmidrule(lr){3-6} \cmidrule(lr){7-9} \cmidrule(lr){10-12}
		& & \begin{tabular}[c]{@{}c@{}}Reasoning\end{tabular} & \begin{tabular}[c]{@{}c@{}}Anomaly\end{tabular} & \begin{tabular}[c]{@{}c@{}}Distribution\end{tabular} & \begin{tabular}[c]{@{}c@{}}Correlation\end{tabular} & \begin{tabular}[c]{@{}c@{}}Range\end{tabular} & \begin{tabular}[c]{@{}c@{}}Order\end{tabular} & \begin{tabular}[c]{@{}c@{}}Filter\end{tabular} & \begin{tabular}[c]{@{}c@{}}Retrieval\end{tabular} & \begin{tabular}[c]{@{}c@{}}Extremum\end{tabular} & \begin{tabular}[c]{@{}c@{}}Cluster\end{tabular} & \\
		\midrule

VisCPM-Chat-v1.1 ~\citep{viscpm}& 10B & 28.4 &\textbf{46.1} & 33.3 & 51.9 & 23.0 & 6.4 & 25.1 & 15.8 & 32.0 & 29.6 & 26.2 \\
BLIP2 ~\citep{li2023blip2}& 11B & 24.8 & 23.4 & 25.0 & 15.1 & 25.3 & 20.2 & 39.8 & 27.8 & 30.3 & 30.1 & 28.3 \\
CogVLM-17B ~\citep{cogvlm}& 17B & 20.3 & 23.1 & 43.6 & 29.6 & 37.7 & 10.8 & 9.1 & 37.9 & 56.6 & 26.7 & 29.4 \\
LLaVA1.5 ~\citep{llava}& 13B & 32.4 & 6.3 & 30.9 & 23.1 & 21.7 & 32.7 & 35.6 & 32.6 & 35.8 & 43.5 & 32.2 \\
ChartAst-S ~\citep{chartassisstant}& 13B & 24.6 & 27.7 & 35.8 & 28.1 & 30.5 & 22.5 & 14.7 & 39.4 & 63.0 & 26.4 & 32.4 \\
MiniCPM-v2 ~\citep{minicpm}& 2.4B & 19.5 & 55.1 & 33.3 & 56.5 & 24.9 & 16.7 & 36.3 & 37.9 & 52.4 & 32.0 & 33.0 \\
mPLUG-Owl2 ~\citep{mplugowl2}& 7B & 31.0 & 27.0 & 29.4 & 35.3 & 28.4 & 22.5 & 40.3 & 30.9 & 41.1 & 27.3 & 33.3 \\
Qwen-VL ~\citep{bai2023qwen}& 7B & 27.8 & 36.3 & 45.1 & 55.8 & 33.8 & 20.0 & 28.7 & 31.3 & 50.2 & 27.1 & 33.4 \\
ViP-LLaVA ~\citep{vipllava}& 13B & 28.8 & 6.6 & 34.8 & 30.3 & 21.9 & \textbf{35.8} & \textbf{40.4} & 42.2 & 38.3 & 33.8 & 33.8 \\
LLaVA-NEXT ~\citep{liu2024llavanext}& 13B & \textbf{30.6} & 7.4 & 26.5 & 38.0 & 29.5 & 33.3 & 23.4 & 53.5 & 59.8 & \textbf{52.3} & 38.5 \\
Sphinx ~\citep{lin2023sphinx}& 13B & 30.0 & 28.9 & 37.8 & 36.1 & 25.8 & 23.5 & 36.7 & 49.7 & \textbf{66.3} & 45.3 & 40.2 \\
\rowcolor{lightblue} 
\textbf{\model(ours)}&  4.6B & 28.6 & 21.5 & \textbf{50.5} & \textbf{58.7} & \textbf{59.5} & 10.4 & 27.3 & \textbf{71.2} & 52.8 & 31.5 & \textbf{42.7} \\
		\bottomrule
	\end{tabular}
}
\end{table}

The ChartInsights benchmark~\citep{chartinsights} evaluates multimodal models' capabilities in low-level chart analysis tasks, challenging them to not only recognize visual elements but also understand their underlying statistical and analytical significance. As shown in Table~\ref{tab:askchart_on_chartinsights}, AskChart demonstrates exceptional performance across various analytical tasks. Notably, it excels in the distribution and correlation tasks, achieving scores of 50\% and 58.7\%, the highest among all evaluated models. Furthermore, AskChart outperforms competitors in the range task with a leading score of 59.5\%. Its performance in retrieval is also remarkable, achieving a score of 71\%, significantly surpassing other models. Overall, AskChart attains an impressive total score of 42.7\%, ranking first among all models. These results highlight the effectiveness of the OCR-aware data prompt strategy employed during pretraining, which has enabled AskChart to align textual and visual semantics effectively, particularly excelling in tasks requiring nuanced integration of both modalities.

\subsection{Ablation Study}

\paragraph{The Impact of Different Prompts.}

\begin{table}[t!]
\begin{center}
\caption{Ablation study on different prompts.} \label{tab:prompt ablation} 
\vspace{-.6em}  
\resizebox{\textwidth}{!}
{ 
\small
\begin{tabular}{ccccccccc}
\toprule
\multirow{2}{*}{Visual Prompt} & \multirow{2}{*}{Ocr-aware data prompt} & \multicolumn{3}{c}{ChartQA} & \multicolumn{1}{c}{Open-ended ChartQA} & \multicolumn{1}{c}{Chart-to-Table} & \multicolumn{2}{c}{Chart-to-Text} \\
\cmidrule(lr){3-5} \cmidrule(lr){6-6} \cmidrule(lr){7-7} \cmidrule(lr){8-9}
 & & aug. & human & avg. & OpenCQA & ChartQA & Pew & Statista  \\
\midrule
\ding{55}    & \ding{55} & 75.5 & 44.9 & 60.2  & 63.1 & 63.9 & 55.2 & 55.1     \\
\ding{55}    & \checkmark & 83.8 & 50.1 & 67.0 & 79.3 & 81.3 & 57.2 & 58.0  \\
\checkmark   & \ding{55} & 76.6 & 46.1 & 61.4  & 63.4 & 62.6 & 50.9 & 55.1  \\
\rowcolor{lightblue} 
\checkmark     & \checkmark & \textbf{84.6} & \textbf{50.9} & \textbf{67.8} & \textbf{79.3} & \textbf{81.5} & \textbf{60.6} & \textbf{62.8}  \\
\bottomrule
\end{tabular}
}
\end{center}
\vspace{-1.5em}
\end{table}

To evaluate the influence of visual prompts and OCR-aware data prompts on model performance, we randomly sampled approximately 1M samples from the sub-datasets of each stage due to limited computational resources. We trained the model from scratch, and the results are shown in Table~\ref{tab:prompt ablation}. The results indicate that visual prompts significantly enhance the model's performance on question-answering tasks (notably, we trained with only about 35\% of the visual prompt dataset). This suggests that visual cues in charts help the model focus on the relevant areas associated with the questions. 

\paragraph{The Impact of Training Strategy.}

To assess which alignment strategy more effectively aligns visual and textual information, we pre-trained the model in Stage I using two different tasks: Chart-to-Text and Chart-to-Table. As shown in Table~\ref{tab:training strategy in stageI}, the model trained with the Chart-to-Table alignment strategy consistently outperforms across multiple tasks. We attribute this to the fact that Chart-to-Table translation helps the model understand the underlying chart content rather than generating potentially irrelevant textual descriptions.

\paragraph{The Impact of Number of Experts.}
To evaluate the effect of the number of experts in the MoE layers on model performance, we conducted the following experiments. First, we varied the total number of experts while keeping the number of activated experts constant. As shown in Table~\ref{tab:total experts ablation}, increasing the number of experts leads to improved performance across various tasks.

\begin{minipage}{0.5\linewidth}
    \centering
    \captionsetup{skip=0pt}
    \includegraphics[width=\linewidth]{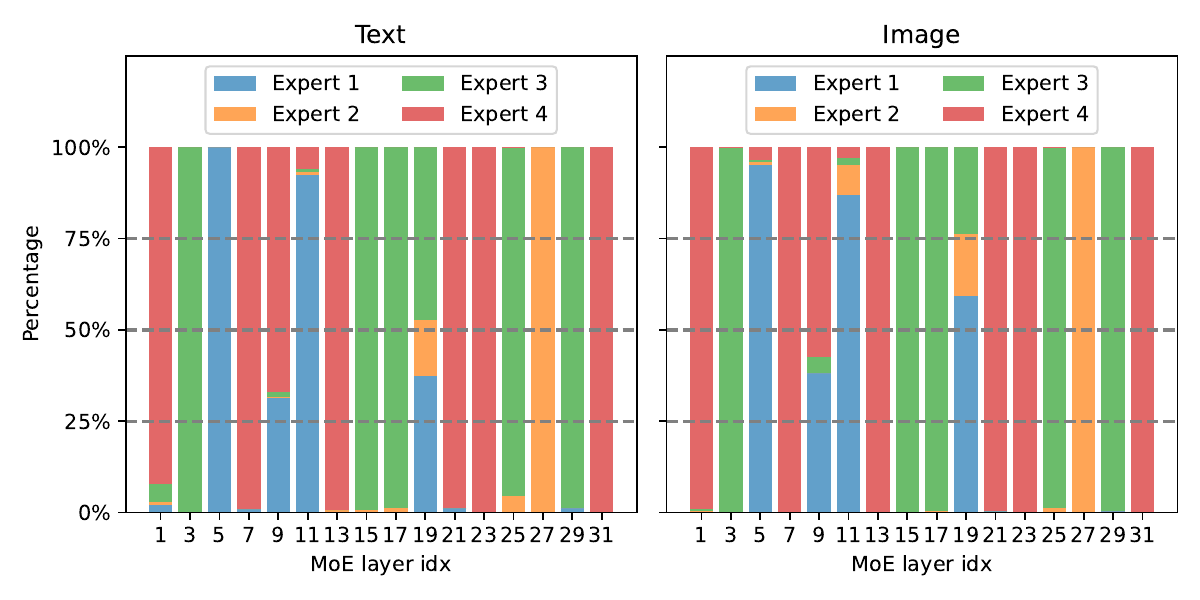}
    \vspace{-1em}
    \captionof{figure}{Modalities across different experts.}
    \label{modalities distribution}
\end{minipage}
\hfill
\begin{minipage}{0.48\linewidth}
    \centering
    \captionsetup{skip=0pt}
    
    \captionof{table}{Ablation study on training stage I.}
    \label{tab:training strategy in stageI}
    \resizebox{\linewidth}{!}{
        \setlength{\tabcolsep}{5pt}
        \begin{tabular}{ccccc}
        \toprule
        Task   &  ChartQA   & Chart-to-Table  & Chart-to-Text \\
        \midrule
        w/ Chart-to-Text    & 67.0  & 80.2 & 59.1    \\
        \rowcolor{lightblue} 
        w/ Chart-to-Table   & \textbf{67.8} & \textbf{81.5} & \textbf{61.7}    \\
        \bottomrule
        \end{tabular}
    }
    \vspace{1em}  
    
    \captionof{table}{Zero-shot study on multiple datasets.}
    \label{tab:zero shot}
    \resizebox{\linewidth}{!}{
        \setlength{\tabcolsep}{5pt}
        \begin{tabular}{ccccc}
        \toprule
        \multirow{2}{*}{Model}   &  \multicolumn{1}{c}{ChartQA}   & \multicolumn{1}{c}{Chart-to-Table}  & \multicolumn{1}{c}{Chart-to-Text} \\
        \cmidrule(lr){2-2} \cmidrule(lr){3-3} \cmidrule(lr){4-4} 
         & RealCQA & StructChart & ChartX  \\
        \midrule
        Unichart~\citep{masry2023unichart}    &  \textbf{38.0}  &  1.6   & 6.8   \\
        LLaVA1.5~\citep{llava}    & 30.0 & 7.5 &  0.45  \\
        LLaVA-NEXT~\citep{liu2024llavanext}    & 33.0 & 14.6 &  14.6  \\
        ChartAst~\citep{chartassisstant}    & 11.0 & 14.3 &  12.8  \\
        \rowcolor{lightblue} 
        AskChart (ours)    & 33.0  & \textbf{30.5}  & \textbf{36.9}   \\
        \bottomrule
        \end{tabular}
    }
\end{minipage}

\begin{table}[thbp]
    \begin{minipage}{0.5\linewidth}
        \centering
        \captionsetup{skip=0pt}
        \caption{The impact of the MoE layers.}
        \resizebox{\linewidth}{!}{
        \setlength{\tabcolsep}{5pt}
        \begin{tabular}{ccccc}
        \toprule
        MoE Layers  &  ChartQA   & Chart-to-Table  & Chart-to-Text \\
        \midrule
        w/o MoE    & 35.9 & 59.1 & 31.2   \\
        \rowcolor{lightblue}
        w/ MoE (\#Experts=4)     & \textbf{76.1} & \textbf{87.4} & \textbf{56.1}   \\
        \bottomrule
        \end{tabular}
        }
        \label{tab:total experts ablation}
    \end{minipage}
    \hfill
    \begin{minipage}{0.48\linewidth}
    \centering
    \captionsetup{skip=0pt}
    \caption{The performance of top-$k$ experts}
    \resizebox{\linewidth}{!}{
    \setlength{\tabcolsep}{5pt}
    \begin{tabular}{ccccc}
    \toprule
    Experts   &  ChartQA   & Chart-to-Table  & Chart-to-Text \\
    \midrule
    1    & 74.4 & 86.4 & 51.5    \\
    \rowcolor{lightblue}
    2    & \textbf{76.1} & \textbf{87.4} & \textbf{56.1}    \\
    \bottomrule
    \end{tabular}
    }
    \label{tab:top-k expert}
    \end{minipage}
\end{table}

Furthermore, as illustrated in Figure~\ref{modalities distribution}, we examined the distribution of different modalities across the experts. Interestingly, the router distribution for both text and image tokens is similar, indicating that each expert is capable of processing both types of tokens. The weighted outputs from multiple experts contribute to stronger multimodal representations.
Next, we varied the number of activated experts while keeping the total number of experts fixed. As presented in Table~\ref{tab:top-k expert}, activating 2 experts yields the best improvement in model performance. To balance computational efficiency and performance, we opted to set the number of activated experts to 2.

\subsection{Zero-shot Study}

To evaluate the generalization capability of our model, we collected data from datasets that the model had never seen before for zero-shot experiments. Specifically, we conducted tests on several datasets, including RealCQA~\citep{ahmed2023realcqa}, StructChart~\citep{xia2023structchart}, and ChartX~\citep{xia2024chartxchartvlmversatile}, for the ChartQA, Chart-to-Table, and Chart-to-Text tasks, respectively. The evaluation metrics were consistent with those used for the corresponding tasks in the main results. As shown in Table~\ref{tab:zero shot}, AskChart exhibited superior zero-shot performance across all tasks.
In contrast, UniChart~\citep{masry2023unichart} performed poorly on both the Chart-to-Table and Chart-to-Text tasks, which we attribute to the limited language modeling capability of its text decoder. Even though ChartAst~\citep{chartassisstant} utilizes a 13B parameter LLM, its generalization ability remains limited. AskChart, with only 4.6B parameters, demonstrated a clear advantage in ChartQA and text generation tasks. It suggests that the text-enhanced visual representation and robust MoE architecture contribute to the model's improved understanding of charts.

\section{Conclusion}
In this paper, we introduced \model, a lightweight chart understanding model that integrates both textual and visual cues using a Mixture of Experts architecture. By employing a three-stage training strategy with tailored pretraining objectives, \model demonstrates enhanced performance across diverse chart understanding tasks. We also presented \dataset, a large-scale dataset with approximately 7.5M samples, featuring three specialized sub-datasets designed to improve the model’s ability to comprehend and interpret chart data. Extensive experiments show that \model achieves state-of-the-art results, outperforming larger models in tasks such as Open-ended ChartQA and Chart-to-Text by 68.3\% and 49.2\%, respectively.



\bibliography{references}

\begin{thebibliography}{63}
\providecommand{\natexlab}[1]{#1}
\providecommand{\url}[1]{\texttt{#1}}
\expandafter\ifx\csname urlstyle\endcsname\relax
  \providecommand{\doi}[1]{doi: #1}\else
  \providecommand{\doi}{doi: \begingroup \urlstyle{rm}\Url}\fi

\bibitem[Pad()]{PaddleOCR}
Paddleocr.
\newblock \url{https://paddlepaddle.github.io/PaddleOCR/}.
\newblock Accessed: 2024-09-30.

\bibitem[Ahmed et~al.(2023)Ahmed, Jawade, Pandey, Setlur, and
  Govindaraju]{ahmed2023realcqa}
Saleem Ahmed, Bhavin Jawade, Shubham Pandey, Srirangaraj Setlur, and Venu
  Govindaraju.
\newblock Realcqa: Scientific chart question answering as a test-bed for
  first-order logic.
\newblock In \emph{International Conference on Document Analysis and
  Recognition}, pp.\  66--83. Springer, 2023.

\bibitem[Bai et~al.(2023)Bai, Bai, Yang, Wang, Tan, Wang, Lin, Zhou, and
  Zhou]{bai2023qwen}
Jinze Bai, Shuai Bai, Shusheng Yang, Shijie Wang, Sinan Tan, Peng Wang, Junyang
  Lin, Chang Zhou, and Jingren Zhou.
\newblock Qwen-vl: A frontier large vision-language model with versatile
  abilities.
\newblock \emph{arXiv preprint arXiv:2308.12966}, 2023.

\bibitem[Cai et~al.(2024{\natexlab{a}})Cai, Liu, Mustikovela, Meyer, Chai,
  Park, and Lee]{cai2024vipllava}
Mu~Cai, Haotian Liu, Siva~Karthik Mustikovela, Gregory~P. Meyer, Yuning Chai,
  Dennis Park, and Yong~Jae Lee.
\newblock Making large multimodal models understand arbitrary visual prompts.
\newblock In \emph{IEEE Conference on Computer Vision and Pattern Recognition},
  2024{\natexlab{a}}.

\bibitem[Cai et~al.(2024{\natexlab{b}})Cai, Liu, Park, Mustikovela, Meyer,
  Chai, and Lee]{vipllava}
Mu~Cai, Haotian Liu, Dennis Park, Siva~Karthik Mustikovela, Gregory~P. Meyer,
  Yuning Chai, and Yong~Jae Lee.
\newblock Vip-llava: Making large multimodal models understand arbitrary visual
  prompts, 2024{\natexlab{b}}.
\newblock URL \url{https://arxiv.org/abs/2312.00784}.

\bibitem[Changpinyo et~al.(2021)Changpinyo, Sharma, Ding, and
  Soricut]{Conceptual}
Soravit Changpinyo, Piyush Sharma, Nan Ding, and Radu Soricut.
\newblock Conceptual 12m: Pushing web-scale image-text pre-training to
  recognize long-tail visual concepts.
\newblock In \emph{Proceedings of the IEEE/CVF conference on computer vision
  and pattern recognition}, pp.\  3558--3568, 2021.

\bibitem[Chen et~al.(2022)Chen, Wang, Changpinyo, Piergiovanni, Padlewski,
  Salz, Goodman, Grycner, Mustafa, Beyer, et~al.]{pali}
Xi~Chen, Xiao Wang, Soravit Changpinyo, AJ~Piergiovanni, Piotr Padlewski,
  Daniel Salz, Sebastian Goodman, Adam Grycner, Basil Mustafa, Lucas Beyer,
  et~al.
\newblock Pali: A jointly-scaled multilingual language-image model.
\newblock \emph{arXiv preprint arXiv:2209.06794}, 2022.

\bibitem[Chiang et~al.(2023)Chiang, Li, Lin, Sheng, Wu, Zhang, Zheng, Zhuang,
  Zhuang, Gonzalez, Stoica, and Xing]{vicuna}
Wei-Lin Chiang, Zhuohan Li, Zi~Lin, Ying Sheng, Zhanghao Wu, Hao Zhang, Lianmin
  Zheng, Siyuan Zhuang, Yonghao Zhuang, Joseph~E. Gonzalez, Ion Stoica, and
  Eric~P. Xing.
\newblock Vicuna: An open-source chatbot impressing gpt-4 with 90\%* chatgpt
  quality.
\newblock March 2023.
\newblock URL \url{https://vicuna.lmsys.org}.

\bibitem[Farin(2014)]{farin2014curves}
Gerald Farin.
\newblock \emph{Curves and surfaces for computer-aided geometric design: a
  practical guide}.
\newblock Elsevier, 2014.

\bibitem[{Fedus} et~al.(2021){Fedus}, {Zoph}, and
  {Shazeer}]{Switch-Transformers}
William {Fedus}, Barret {Zoph}, and Noam {Shazeer}.
\newblock {Switch Transformers: Scaling to Trillion Parameter Models with
  Simple and Efficient Sparsity}.
\newblock \emph{arXiv e-prints}, art. arXiv:2101.03961, January 2021.
\newblock \doi{10.48550/arXiv.2101.03961}.

\bibitem[Han et~al.(2023)Han, Zhang, Chen, Yang, Wang, Yu, Fu, and
  Zhang]{chartllama}
Yucheng Han, Chi Zhang, Xin Chen, Xu~Yang, Zhibin Wang, Gang Yu, Bin Fu, and
  Hanwang Zhang.
\newblock Chartllama: A multimodal llm for chart understanding and generation.
\newblock \emph{arXiv preprint arXiv:2311.16483}, 2023.

\bibitem[Hoque et~al.(2022)Hoque, Kavehzadeh, and
  Masry]{chart_question_answering}
Enamul Hoque, Parsa Kavehzadeh, and Ahmed Masry.
\newblock Chart question answering: State of the art and future directions.
\newblock In \emph{Computer Graphics Forum}, volume~41, pp.\  555--572. Wiley
  Online Library, 2022.

\bibitem[Hu et~al.(2023)Hu, Yao, Wang, Wang, Pan, Chen, Yu, Wu, Zhao, Zhang,
  Han, Lin, Xue, Li, Liu, and Sun]{viscpm}
Jinyi Hu, Yuan Yao, Chongyi Wang, Shan Wang, Yinxu Pan, Qianyu Chen, Tianyu Yu,
  Hanghao Wu, Yue Zhao, Haoye Zhang, Xu~Han, Yankai Lin, Jiao Xue, Dahai Li,
  Zhiyuan Liu, and Maosong Sun.
\newblock Large multilingual models pivot zero-shot multimodal learning across
  languages.
\newblock \emph{arXiv preprint arXiv:2308.12038}, 2023.

\bibitem[Hu et~al.(2024)Hu, Tu, Han, He, Cui, Long, Zheng, Fang, Huang, Zhao,
  Zhang, Thai, Zhang, Wang, Yao, Zhao, Zhou, Cai, Zhai, Ding, Jia, Zeng, Li,
  Liu, and Sun]{minicpm}
Shengding Hu, Yuge Tu, Xu~Han, Chaoqun He, Ganqu Cui, Xiang Long, Zhi Zheng,
  Yewei Fang, Yuxiang Huang, Weilin Zhao, Xinrong Zhang, Zheng~Leng Thai,
  Kaihuo Zhang, Chongyi Wang, Yuan Yao, Chenyang Zhao, Jie Zhou, Jie Cai,
  Zhongwu Zhai, Ning Ding, Chao Jia, Guoyang Zeng, Dahai Li, Zhiyuan Liu, and
  Maosong Sun.
\newblock Minicpm: Unveiling the potential of small language models with
  scalable training strategies, 2024.
\newblock URL \url{https://arxiv.org/abs/2404.06395}.

\bibitem[Huang et~al.(2024)Huang, Chan, Fung, Qiu, Zhou, Joty, Chang, and
  Ji]{huang2024pixelsinsightssurveyautomatic}
Kung-Hsiang Huang, Hou~Pong Chan, Yi~R. Fung, Haoyi Qiu, Mingyang Zhou, Shafiq
  Joty, Shih-Fu Chang, and Heng Ji.
\newblock From pixels to insights: A survey on automatic chart understanding in
  the era of large foundation models, 2024.
\newblock URL \url{https://arxiv.org/abs/2403.12027}.

\bibitem[Kantharaj et~al.(2022{\natexlab{a}})Kantharaj, Do, Leong, Tan, Hoque,
  and Joty]{kantharaj2022opencqa}
Shankar Kantharaj, Xuan~Long Do, Rixie~Tiffany Leong, Jia~Qing Tan, Enamul
  Hoque, and Shafiq Joty.
\newblock {O}pen{CQA}: Open-ended question answering with charts.
\newblock In \emph{Proceedings of the 2022 Conference on Empirical Methods in
  Natural Language Processing}, pp.\  11817--11837, Abu Dhabi, United Arab
  Emirates, December 2022{\natexlab{a}}. Association for Computational
  Linguistics.
\newblock URL \url{https://aclanthology.org/2022.emnlp-main.811}.

\bibitem[Kantharaj et~al.(2022{\natexlab{b}})Kantharaj, Leong, Lin, Masry,
  Thakkar, Hoque, and Joty]{kantharaj2022charttotext}
Shankar Kantharaj, Rixie~Tiffany Leong, Xiang Lin, Ahmed Masry, Megh Thakkar,
  Enamul Hoque, and Shafiq Joty.
\newblock Chart-to-text: A large-scale benchmark for chart summarization.
\newblock In Smaranda Muresan, Preslav Nakov, and Aline Villavicencio (eds.),
  \emph{Proceedings of the 60th Annual Meeting of the Association for
  Computational Linguistics (Volume 1: Long Papers)}, pp.\  4005--4023, Dublin,
  Ireland, May 2022{\natexlab{b}}. Association for Computational Linguistics.
\newblock \doi{10.18653/v1/2022.acl-long.277}.
\newblock URL \url{https://aclanthology.org/2022.acl-long.277}.

\bibitem[{Kil} et~al.(2022){Kil}, {Changpinyo}, {Chen}, {Hu}, {Goodman},
  {Chao}, and {Soricut}]{PreSTU}
Jihyung {Kil}, Soravit {Changpinyo}, Xi~{Chen}, Hexiang {Hu}, Sebastian
  {Goodman}, Wei-Lun {Chao}, and Radu {Soricut}.
\newblock {PreSTU: Pre-Training for Scene-Text Understanding}.
\newblock \emph{arXiv e-prints}, art. arXiv:2209.05534, September 2022.
\newblock \doi{10.48550/arXiv.2209.05534}.

\bibitem[Kim et~al.(2022)Kim, Hong, Yim, Nam, Park, Yim, Hwang, Yun, Han, and
  Park]{Donut}
Geewook Kim, Teakgyu Hong, Moonbin Yim, JeongYeon Nam, Jinyoung Park, Jinyeong
  Yim, Wonseok Hwang, Sangdoo Yun, Dongyoon Han, and Seunghyun Park.
\newblock Ocr-free document understanding transformer.
\newblock In \emph{European Conference on Computer Vision}, pp.\  498--517.
  Springer, 2022.

\bibitem[Lee et~al.(2023)Lee, Joshi, Turc, Hu, Liu, Eisenschlos, Khandelwal,
  Shaw, Chang, and Toutanova]{pix2struct}
Kenton Lee, Mandar Joshi, Iulia~Raluca Turc, Hexiang Hu, Fangyu Liu,
  Julian~Martin Eisenschlos, Urvashi Khandelwal, Peter Shaw, Ming-Wei Chang,
  and Kristina Toutanova.
\newblock Pix2struct: Screenshot parsing as pretraining for visual language
  understanding.
\newblock In \emph{International Conference on Machine Learning}, pp.\
  18893--18912. PMLR, 2023.

\bibitem[Li et~al.(2024)Li, Hui, Qu, Yang, Li, Li, Wang, Qin, Geng, Huo, Zhou,
  Ma, Li, Chang, Huang, Cheng, and Li]{bird}
Jinyang Li, Binyuan Hui, Ge~Qu, Jiaxi Yang, Binhua Li, Bowen Li, Bailin Wang,
  Bowen Qin, Ruiying Geng, Nan Huo, Xuanhe Zhou, Chenhao Ma, Guoliang Li,
  Kevin~C.C. Chang, Fei Huang, Reynold Cheng, and Yongbin Li.
\newblock Can llm already serve as a database interface? a big bench for
  large-scale database grounded text-to-sqls.
\newblock In \emph{Proceedings of the 37th International Conference on Neural
  Information Processing Systems}, NIPS '23, Red Hook, NY, USA, 2024. Curran
  Associates Inc.

\bibitem[Li et~al.(2023{\natexlab{a}})Li, Li, Savarese, and Hoi]{li2023blip2}
Junnan Li, Dongxu Li, Silvio Savarese, and Steven Hoi.
\newblock Blip-2: Bootstrapping language-image pre-training with frozen image
  encoders and large language models.
\newblock In \emph{International conference on machine learning}, pp.\
  19730--19742. PMLR, 2023{\natexlab{a}}.

\bibitem[Li et~al.(2023{\natexlab{b}})Li, Bubeck, Eldan, Del~Giorno, Gunasekar,
  and Lee]{phi}
Yuanzhi Li, S{\'e}bastien Bubeck, Ronen Eldan, Allie Del~Giorno, Suriya
  Gunasekar, and Yin~Tat Lee.
\newblock Textbooks are all you need ii: phi-1.5 technical report.
\newblock \emph{arXiv preprint arXiv:2309.05463}, 2023{\natexlab{b}}.

\bibitem[Lin et~al.(2024)Lin, Tang, Ye, Cui, Zhu, Jin, Zhang, Ning, and
  Yuan]{moe-llava}
Bin Lin, Zhenyu Tang, Yang Ye, Jiaxi Cui, Bin Zhu, Peng Jin, Junwu Zhang, Munan
  Ning, and Li~Yuan.
\newblock Moe-llava: Mixture of experts for large vision-language models.
\newblock \emph{arXiv preprint arXiv:2401.15947}, 2024.

\bibitem[Lin et~al.(2014)Lin, Maire, Belongie, Hays, Perona, Ramanan,
  Doll{\'a}r, and Zitnick]{coco}
Tsung-Yi Lin, Michael Maire, Serge Belongie, James Hays, Pietro Perona, Deva
  Ramanan, Piotr Doll{\'a}r, and C~Lawrence Zitnick.
\newblock Microsoft coco: Common objects in context.
\newblock In \emph{Computer Vision--ECCV 2014: 13th European Conference,
  Zurich, Switzerland, September 6-12, 2014, Proceedings, Part V 13}, pp.\
  740--755. Springer, 2014.

\bibitem[Lin et~al.(2023)Lin, Liu, Zhang, Gao, Qiu, Xiao, Qiu, Lin, Shao, Chen,
  et~al.]{lin2023sphinx}
Ziyi Lin, Chris Liu, Renrui Zhang, Peng Gao, Longtian Qiu, Han Xiao, Han Qiu,
  Chen Lin, Wenqi Shao, Keqin Chen, et~al.
\newblock Sphinx: The joint mixing of weights, tasks, and visual embeddings for
  multi-modal large language models.
\newblock \emph{arXiv preprint arXiv:2311.07575}, 2023.

\bibitem[Liu et~al.(2022)Liu, Piccinno, Krichene, Pang, Lee, Joshi, Altun,
  Collier, and Eisenschlos]{liu2022matcha}
Fangyu Liu, Francesco Piccinno, Syrine Krichene, Chenxi Pang, Kenton Lee,
  Mandar Joshi, Yasemin Altun, Nigel Collier, and Julian~Martin Eisenschlos.
\newblock Matcha: Enhancing visual language pretraining with math reasoning and
  chart derendering.
\newblock \emph{arXiv preprint arXiv:2212.09662}, 2022.

\bibitem[Liu et~al.(2023)Liu, Eisenschlos, Piccinno, Krichene, Pang, Lee,
  Joshi, Chen, Collier, and Altun]{deplot}
Fangyu Liu, Julian Eisenschlos, Francesco Piccinno, Syrine Krichene, Chenxi
  Pang, Kenton Lee, Mandar Joshi, Wenhu Chen, Nigel Collier, and Yasemin Altun.
\newblock {D}e{P}lot: One-shot visual language reasoning by plot-to-table
  translation.
\newblock In Anna Rogers, Jordan Boyd-Graber, and Naoaki Okazaki (eds.),
  \emph{Findings of the Association for Computational Linguistics: ACL 2023},
  pp.\  10381--10399, Toronto, Canada, July 2023. Association for Computational
  Linguistics.

\bibitem[Liu et~al.(2024{\natexlab{a}})Liu, Li, Li, Li, Zhang, Shen, and
  Lee]{liu2024llavanext}
Haotian Liu, Chunyuan Li, Yuheng Li, Bo~Li, Yuanhan Zhang, Sheng Shen, and
  Yong~Jae Lee.
\newblock Llava-next: Improved reasoning, ocr, and world knowledge, January
  2024{\natexlab{a}}.
\newblock URL \url{https://llava-vl.github.io/blog/2024-01-30-llava-next/}.

\bibitem[Liu et~al.(2024{\natexlab{b}})Liu, Li, Wu, and Lee]{llava}
Haotian Liu, Chunyuan Li, Qingyang Wu, and Yong~Jae Lee.
\newblock Visual instruction tuning.
\newblock \emph{Advances in neural information processing systems}, 36,
  2024{\natexlab{b}}.

\bibitem[Luo et~al.(2018)Luo, Qin, Tang, and Li]{luo2018deepeye}
Yuyu Luo, Xuedi Qin, Nan Tang, and Guoliang Li.
\newblock Deepeye: Towards automatic data visualization.
\newblock In \emph{2018 IEEE 34th international conference on data engineering
  (ICDE)}, pp.\  101--112. IEEE, 2018.

\bibitem[Luo et~al.(2021)Luo, Tang, Li, Chai, Li, and
  Qin]{DBLP:conf/sigmod/Luo00CLQ21}
Yuyu Luo, Nan Tang, Guoliang Li, Chengliang Chai, Wenbo Li, and Xuedi Qin.
\newblock Synthesizing natural language to visualization {(NL2VIS)} benchmarks
  from {NL2SQL} benchmarks.
\newblock In \emph{{SIGMOD} Conference}, pp.\  1235--1247. {ACM}, 2021.

\bibitem[Luo et~al.(2022{\natexlab{a}})Luo, Qin, Chai, Tang, Li, and
  Li]{DBLP:journals/tkde/LuoQCTLL22}
Yuyu Luo, Xuedi Qin, Chengliang Chai, Nan Tang, Guoliang Li, and Wenbo Li.
\newblock Steerable self-driving data visualization.
\newblock \emph{{IEEE} Trans. Knowl. Data Eng.}, 34\penalty0 (1):\penalty0
  475--490, 2022{\natexlab{a}}.

\bibitem[Luo et~al.(2022{\natexlab{b}})Luo, Tang, Li, Tang, Chai, and
  Qin]{DBLP:journals/tvcg/LuoTLTCQ22}
Yuyu Luo, Nan Tang, Guoliang Li, Jiawei Tang, Chengliang Chai, and Xuedi Qin.
\newblock Natural language to visualization by neural machine translation.
\newblock \emph{{IEEE} Trans. Vis. Comput. Graph.}, 28\penalty0 (1):\penalty0
  217--226, 2022{\natexlab{b}}.

\bibitem[Luo et~al.(2023)Luo, Zhou, Tang, Li, Chai, and
  Shen]{DBLP:journals/pacmmod/LuoZ00CS23}
Yuyu Luo, Yihui Zhou, Nan Tang, Guoliang Li, Chengliang Chai, and Leixian Shen.
\newblock Learned data-aware image representations of line charts for
  similarity search.
\newblock \emph{Proc. {ACM} Manag. Data}, 1\penalty0 (1):\penalty0 88:1--88:29,
  2023.

\bibitem[Masry et~al.(2022)Masry, Do, Tan, Joty, and Hoque]{masry2022chartqa}
Ahmed Masry, Xuan~Long Do, Jia~Qing Tan, Shafiq Joty, and Enamul Hoque.
\newblock {C}hart{QA}: A benchmark for question answering about charts with
  visual and logical reasoning.
\newblock In Smaranda Muresan, Preslav Nakov, and Aline Villavicencio (eds.),
  \emph{Findings of the Association for Computational Linguistics: ACL 2022},
  pp.\  2263--2279, Dublin, Ireland, May 2022. Association for Computational
  Linguistics.
\newblock \doi{10.18653/v1/2022.findings-acl.177}.
\newblock URL \url{https://aclanthology.org/2022.findings-acl.177}.

\bibitem[Masry et~al.(2023)Masry, Kavehzadeh, Do, Hoque, and
  Joty]{masry2023unichart}
Ahmed Masry, Parsa Kavehzadeh, Xuan~Long Do, Enamul Hoque, and Shafiq Joty.
\newblock {U}ni{C}hart: A universal vision-language pretrained model for chart
  comprehension and reasoning.
\newblock In Houda Bouamor, Juan Pino, and Kalika Bali (eds.),
  \emph{Proceedings of the 2023 Conference on Empirical Methods in Natural
  Language Processing}, pp.\  14662--14684, Singapore, December 2023.
  Association for Computational Linguistics.
\newblock \doi{10.18653/v1/2023.emnlp-main.906}.
\newblock URL \url{https://aclanthology.org/2023.emnlp-main.906}.

\bibitem[Masry et~al.(2024{\natexlab{a}})Masry, Shahmohammadi, Parvez, Hoque,
  and Joty]{masry2024chartinstruct}
Ahmed Masry, Mehrad Shahmohammadi, Md~Rizwan Parvez, Enamul Hoque, and Shafiq
  Joty.
\newblock Chartinstruct: Instruction tuning for chart comprehension and
  reasoning.
\newblock \emph{arXiv preprint arXiv:2403.09028}, 2024{\natexlab{a}}.

\bibitem[Masry et~al.(2024{\natexlab{b}})Masry, Shahmohammadi, Parvez, Hoque,
  and Joty]{masry2024chartinstructinstructiontuningchart}
Ahmed Masry, Mehrad Shahmohammadi, Md~Rizwan Parvez, Enamul Hoque, and Shafiq
  Joty.
\newblock Chartinstruct: Instruction tuning for chart comprehension and
  reasoning, 2024{\natexlab{b}}.
\newblock URL \url{https://arxiv.org/abs/2403.09028}.

\bibitem[Masry et~al.(2024{\natexlab{c}})Masry, Thakkar, Bajaj, Kartha, Hoque,
  and Joty]{masry2024chartgemmavisualinstructiontuningchart}
Ahmed Masry, Megh Thakkar, Aayush Bajaj, Aaryaman Kartha, Enamul Hoque, and
  Shafiq Joty.
\newblock Chartgemma: Visual instruction-tuning for chart reasoning in the
  wild, 2024{\natexlab{c}}.
\newblock URL \url{https://arxiv.org/abs/2407.04172}.

\bibitem[Meng et~al.(2024)Meng, Shao, Lu, Gao, Zhang, Qiao, and
  Luo]{chartassisstant}
Fanqing Meng, Wenqi Shao, Quanfeng Lu, Peng Gao, Kaipeng Zhang, Yu~Qiao, and
  Ping Luo.
\newblock Chartassisstant: A universal chart multimodal language model via
  chart-to-table pre-training and multitask instruction tuning.
\newblock \emph{arXiv preprint arXiv:2401.02384}, 2024.

\bibitem[Methani et~al.(2020)Methani, Ganguly, Khapra, and
  Kumar]{methani2020plotqa}
Nitesh Methani, Pritha Ganguly, Mitesh~M. Khapra, and Pratyush Kumar.
\newblock Plotqa: Reasoning over scientific plots.
\newblock In \emph{The IEEE Winter Conference on Applications of Computer
  Vision (WACV)}, March 2020.

\bibitem[Ouyang et~al.(2022)Ouyang, Wu, Jiang, Almeida, Wainwright, Mishkin,
  Zhang, Agarwal, Slama, Gray, Schulman, Hilton, Kelton, Miller, Simens,
  Askell, Welinder, Christiano, Leike, and Lowe]{ouyang2022training}
Long Ouyang, Jeffrey Wu, Xu~Jiang, Diogo Almeida, Carroll Wainwright, Pamela
  Mishkin, Chong Zhang, Sandhini Agarwal, Katarina Slama, Alex Gray, John
  Schulman, Jacob Hilton, Fraser Kelton, Luke Miller, Maddie Simens, Amanda
  Askell, Peter Welinder, Paul Christiano, Jan Leike, and Ryan Lowe.
\newblock Training language models to follow instructions with human feedback.
\newblock In Alice~H. Oh, Alekh Agarwal, Danielle Belgrave, and Kyunghyun Cho
  (eds.), \emph{Advances in Neural Information Processing Systems}, 2022.
\newblock URL \url{https://openreview.net/forum?id=TG8KACxEON}.

\bibitem[Post(2018)]{2018bleu}
Matt Post.
\newblock A call for clarity in reporting bleu scores.
\newblock \emph{arXiv preprint arXiv:1804.08771}, 2018.

\bibitem[Qin et~al.(2020)Qin, Luo, Tang, and Li]{DBLP:journals/vldb/QinLTL20}
Xuedi Qin, Yuyu Luo, Nan Tang, and Guoliang Li.
\newblock Making data visualization more efficient and effective: a survey.
\newblock \emph{{VLDB} J.}, 29\penalty0 (1):\penalty0 93--117, 2020.

\bibitem[Raffel et~al.(2020)Raffel, Shazeer, Roberts, Lee, Narang, Matena,
  Zhou, Li, and Liu]{T5}
Colin Raffel, Noam Shazeer, Adam Roberts, Katherine Lee, Sharan Narang, Michael
  Matena, Yanqi Zhou, Wei Li, and Peter~J Liu.
\newblock Exploring the limits of transfer learning with a unified text-to-text
  transformer.
\newblock \emph{Journal of machine learning research}, 21\penalty0
  (140):\penalty0 1--67, 2020.

\bibitem[Ren et~al.(2016)Ren, He, Girshick, and Sun]{ren2016faster_R-CNN}
Shaoqing Ren, Kaiming He, Ross Girshick, and Jian Sun.
\newblock Faster r-cnn: Towards real-time object detection with region proposal
  networks.
\newblock \emph{IEEE transactions on pattern analysis and machine
  intelligence}, 39\penalty0 (6):\penalty0 1137--1149, 2016.

\bibitem[Saket et~al.(2019)Saket, Endert, and
  Demiralp]{DBLP:journals/tvcg/SaketED19}
Bahador Saket, Alex Endert, and {\c{C}}agatay Demiralp.
\newblock Task-based effectiveness of basic visualizations.
\newblock \emph{{IEEE} Trans. Vis. Comput. Graph.}, 25\penalty0 (7):\penalty0
  2505--2512, 2019.

\bibitem[Tang et~al.(2022)Tang, Luo, Ouzzani, Li, and
  Chen]{DBLP:conf/sigmod/TangLOLC22}
Jiawei Tang, Yuyu Luo, Mourad Ouzzani, Guoliang Li, and Hongyang Chen.
\newblock Sevi: Speech-to-visualization through neural machine translation.
\newblock In \emph{{SIGMOD} Conference}, pp.\  2353--2356. {ACM}, 2022.

\bibitem[Touvron et~al.(2023)Touvron, Martin, Stone, Albert, Almahairi, Babaei,
  Bashlykov, Batra, Bhargava, Bhosale, et~al.]{touvron2023llama}
Hugo Touvron, Louis Martin, Kevin Stone, Peter Albert, Amjad Almahairi, Yasmine
  Babaei, Nikolay Bashlykov, Soumya Batra, Prajjwal Bhargava, Shruti Bhosale,
  et~al.
\newblock Llama 2: Open foundation and fine-tuned chat models.
\newblock \emph{arXiv preprint arXiv:2307.09288}, 2023.

\bibitem[Wang et~al.(2024)Wang, Lv, Yu, Hong, Qi, Wang, Ji, Yang, Zhao, Song,
  Xu, Xu, Li, Dong, Ding, and Tang]{cogvlm}
Weihan Wang, Qingsong Lv, Wenmeng Yu, Wenyi Hong, Ji~Qi, Yan Wang, Junhui Ji,
  Zhuoyi Yang, Lei Zhao, Xixuan Song, Jiazheng Xu, Bin Xu, Juanzi Li, Yuxiao
  Dong, Ming Ding, and Jie Tang.
\newblock Cogvlm: Visual expert for pretrained language models, 2024.
\newblock URL \url{https://arxiv.org/abs/2311.03079}.

\bibitem[Wei et~al.(2022)Wei, Wang, Schuurmans, Bosma, Xia, Chi, Le, Zhou,
  et~al.]{wei2022Chain-of-thought}
Jason Wei, Xuezhi Wang, Dale Schuurmans, Maarten Bosma, Fei Xia, Ed~Chi, Quoc~V
  Le, Denny Zhou, et~al.
\newblock Chain-of-thought prompting elicits reasoning in large language
  models.
\newblock \emph{Advances in neural information processing systems},
  35:\penalty0 24824--24837, 2022.

\bibitem[Wu et~al.(2024)Wu, Yan, Shen, Wang, Tang, and Luo]{chartinsights}
Yifan Wu, Lutao Yan, Leixian Shen, Yunhai Wang, Nan Tang, and Yuyu Luo.
\newblock Chartinsights: Evaluating multimodal large language models for
  low-level chart question answering.
\newblock In \emph{{EMNLP} (Findings)}. Association for Computational
  Linguistics, 2024.

\bibitem[Xia et~al.(2023)Xia, Zhang, Peng, Ye, Yan, Ye, Shi, Qiao, and
  Yan]{xia2023structchart}
Renqiu Xia, Bo~Zhang, Haoyang Peng, Hancheng Ye, Xiangchao Yan, Peng Ye, Botian
  Shi, Yu~Qiao, and Junchi Yan.
\newblock Structchart: Perception, structuring, reasoning for visual chart
  understanding.
\newblock \emph{arXiv preprint arXiv:2309.11268}, 2023.

\bibitem[Xia et~al.(2024)Xia, Zhang, Ye, Yan, Liu, Zhou, Chen, Dou, Shi, Yan,
  and Qiao]{xia2024chartxchartvlmversatile}
Renqiu Xia, Bo~Zhang, Hancheng Ye, Xiangchao Yan, Qi~Liu, Hongbin Zhou, Zijun
  Chen, Min Dou, Botian Shi, Junchi Yan, and Yu~Qiao.
\newblock Chartx \& chartvlm: A versatile benchmark and foundation model for
  complicated chart reasoning, 2024.
\newblock URL \url{https://arxiv.org/abs/2402.12185}.

\bibitem[Xu et~al.(2020)Xu, Li, Cui, Huang, Wei, and Zhou]{xu2020layoutlm}
Yiheng Xu, Minghao Li, Lei Cui, Shaohan Huang, Furu Wei, and Ming Zhou.
\newblock Layoutlm: Pre-training of text and layout for document image
  understanding.
\newblock In \emph{Proceedings of the 26th ACM SIGKDD international conference
  on knowledge discovery \& data mining}, pp.\  1192--1200, 2020.

\bibitem[Xu et~al.(2023)Xu, Du, Qi, Xu, Yuan, and Guo]{xu2023chartbench}
Zhengzhuo Xu, Sinan Du, Yiyan Qi, Chengjin Xu, Chun Yuan, and Jian Guo.
\newblock Chartbench: A benchmark for complex visual reasoning in charts.
\newblock \emph{arXiv preprint arXiv:2312.15915}, 2023.

\bibitem[Xu et~al.(2024)Xu, Qu, Qi, Du, Xu, Yuan, and
  Guo]{xu2024chartmoemixtureexpertconnector}
Zhengzhuo Xu, Bowen Qu, Yiyan Qi, Sinan Du, Chengjin Xu, Chun Yuan, and Jian
  Guo.
\newblock Chartmoe: Mixture of expert connector for advanced chart
  understanding, 2024.
\newblock URL \url{https://arxiv.org/abs/2409.03277}.

\bibitem[Ye et~al.(2023)Ye, Xu, Ye, Yan, Hu, Liu, Qian, Zhang, Huang, and
  Zhou]{mplugowl2}
Qinghao Ye, Haiyang Xu, Jiabo Ye, Ming Yan, Anwen Hu, Haowei Liu, Qi~Qian,
  Ji~Zhang, Fei Huang, and Jingren Zhou.
\newblock mplug-owl2: Revolutionizing multi-modal large language model with
  modality collaboration, 2023.
\newblock URL \url{https://arxiv.org/abs/2311.04257}.

\bibitem[Yu et~al.(2018)Yu, Zhang, Yang, Yasunaga, Wang, Li, Ma, Li, Yao,
  Roman, Zhang, and Radev]{spider}
Tao Yu, Rui Zhang, Kai Yang, Michihiro Yasunaga, Dongxu Wang, Zifan Li, James
  Ma, Irene Li, Qingning Yao, Shanelle Roman, Zilin Zhang, and Dragomir Radev.
\newblock Spider: A large-scale human-labeled dataset for complex and
  cross-domain semantic parsing and text-to-sql task.
\newblock In \emph{Proceedings of the 2018 Conference on Empirical Methods in
  Natural Language Processing}, Brussels, Belgium, 2018. Association for
  Computational Linguistics.

\bibitem[Zhai et~al.(2023)Zhai, Mustafa, Kolesnikov, and Beyer]{siglip}
Xiaohua Zhai, Basil Mustafa, Alexander Kolesnikov, and Lucas Beyer.
\newblock Sigmoid loss for language image pre-training.
\newblock In \emph{Proceedings of the IEEE/CVF International Conference on
  Computer Vision}, pp.\  11975--11986, 2023.

\bibitem[Zhang et~al.(2023)Zhang, Zhang, Gu, Zhou, Lipka, Yang, and
  Sun]{llavar}
Yanzhe Zhang, Ruiyi Zhang, Jiuxiang Gu, Yufan Zhou, Nedim Lipka, Diyi Yang, and
  Tong Sun.
\newblock Llavar: Enhanced visual instruction tuning for text-rich image
  understanding.
\newblock \emph{arXiv preprint arXiv:2306.17107}, 2023.

\bibitem[Zhou et~al.(2023)Zhou, Fung, Chen, Thomas, Ji, and Chang]{chart-t5}
Mingyang Zhou, Yi~R Fung, Long Chen, Christopher Thomas, Heng Ji, and Shih-Fu
  Chang.
\newblock Enhanced chart understanding in vision and language task via
  cross-modal pre-training on plot table pairs.
\newblock \emph{arXiv preprint arXiv:2305.18641}, 2023.

\end{thebibliography}
\bibliographystyle{iclr2025_conference}

\newpage
\appendix

\section{Statistics of \dataset}
\label{sta_chartbank}
\begin{table}[th]
\begin{center}
	\vspace{-1em}
\caption{Statistics of \dataset dataset.} 
\label{tab:stat_chartbase} 
\vspace{-.5em}
{ \small
\begin{tabular}{lcccc}
\toprule
\textbf{Subdatasets in \dataset}   & \textbf{\#-Chart Types}  & \textbf{\#-Charts} & \textbf{\#-Samples} \\
\midrule
Visual Prompt    &6  &104,445  &417,780   \\
OCR-aware Data Prompt &6 &505,037 &6,980,977 \\
 \ \ - Single-turn    &6  &505,037   &6,791,230  \\
 \ \ - Multi-turn    &6  &189,747   &189,747  \\
Chart-to-Table Instr. Following    &7  &61,472  &61,472   \\
\bottomrule
\end{tabular}
}
\end{center}
\end{table}

\begin{table}[th]
	\small
	\vspace{-1.em}
	\centering 
	\caption{Tasks and datasets used for pretraining (Stage I and Stage II) and fine-tuning (Stage III). 
			Our proposed dataset is denoted by ``\#''.  ``*'' indicates that only a subset of the dataset is used for the task. All datasets are accompanied by data prompts, except the visual prompt dataset.} 
    \label{tab:pretrain_dataset}
    \vspace{-0.5em}
    \resizebox{0.7\textwidth}{!}{
		\begin{tabular}{llll}
			\toprule
			& \textbf{Tasks} & \textbf{Datasets} & \textbf{\#-Samples} \\
			\midrule
			Stage I & Chart-to-Table        & $^\#$OCR-aware Data Prompt         & 495K \\
			\midrule
			\multirow{4}{*}{Stage II} & Chart Summarization  & $^\#$OCR-aware Data Prompt          & 481K \\
			& Num \& Vis Reasoning & $^\#$OCR-aware Data Prompt        & 5.5M \\
			& Open-ended ChartQA     & $^\#$OCR-aware Data Prompt              & 481K \\
			& Low-level ChartQA         & $^\#$Visual Prompt         & 418K \\
			\midrule        
			\multirow{4}{*}{Stage III}& Chart-to-Text   & Chart-to-Text~\citep{kantharaj2022charttotext}         & 35K \\
			& Open-ended ChartQA    & OpenCQA~\citep{kantharaj2022opencqa}               & 5K \\
			& \multirow{2}{*}{Chart-to-Table}   & $^*$ChartQA~\citep{masry2022chartqa}              & 28K \\
			&                                    & $^\#$Chart-to-Table Instruction-Following     & 61K \\
			& Chart QA         & $^*$ChartQA~\citep{masry2022chartqa}              & 28K \\
			\bottomrule
		\end{tabular}
	}
\end{table}

\vspace{-1.5em}
\section{Additional Results from Evaluation}
\subsection{Error Analysis}
\label{error-analysis}

\begin{figure*}[htbp]
    \centering
    \begin{minipage}{0.48\linewidth}
        \centering
        \includegraphics[width=\linewidth]{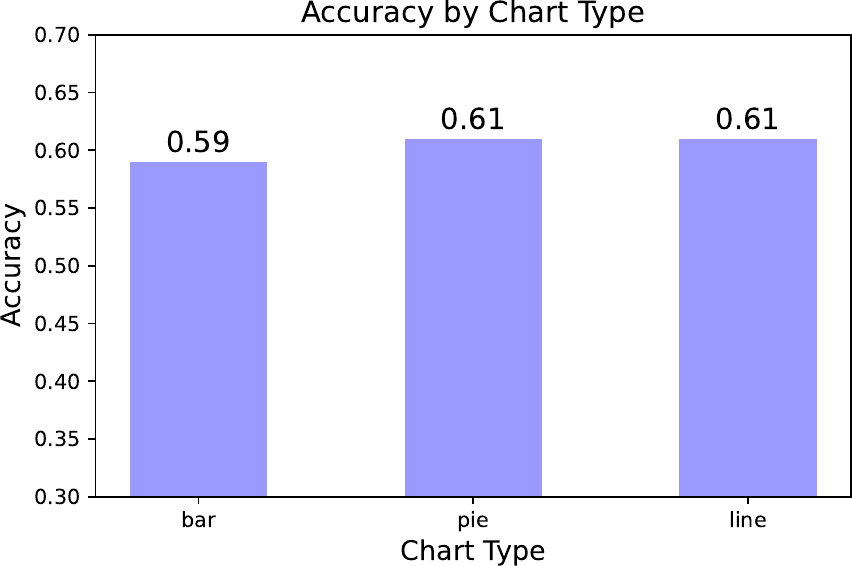}
        \vspace{-1.5em}
        \caption{Results on the ChartQA Human test set by chart type.}
        \label{fig:chart_type_accuracies}
    \end{minipage}
    \hspace{0.01\linewidth} 
    \begin{minipage}{0.48\linewidth}
        \centering
        \includegraphics[width=\linewidth]{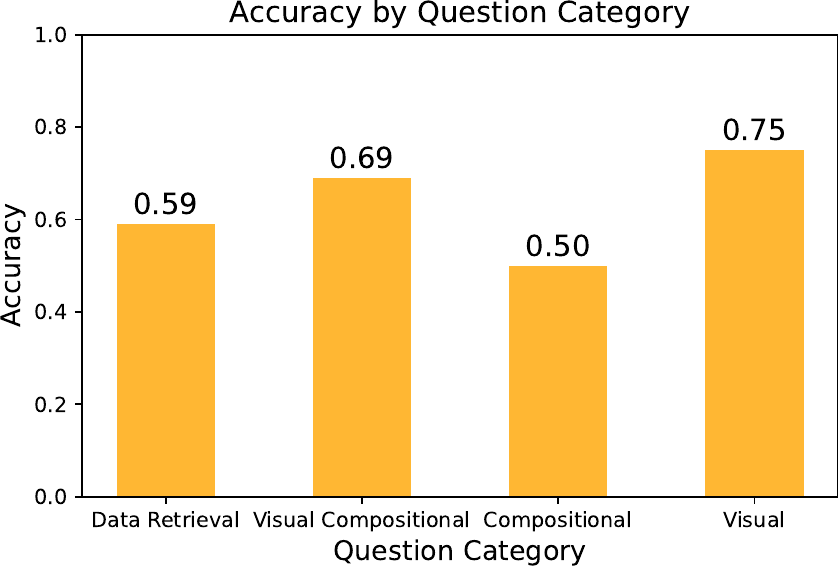}
        \vspace{-1.5em}
        \caption{Results on the ChartQA Human test set by question type.}
        \label{fig:category_accuracies}
    \end{minipage}
\end{figure*}

Figure~\ref{fig:chart_type_accuracies} presents the results across different chart types on the ChartQA-H benchmark. We randomly selected 1,108 human-written questions for this analysis. Figure~\ref{fig:category_accuracies} shows the performance breakdown by question type on ChartQA-H. The question types are as follows: (1) Data Retrieval: Questions focusing on directly extracting data information from the chart; (2) Visual Compositional: Tasks that involve identifying visual elements followed by reasoning to derive an answer; (3) Compositional: Multi-step reasoning or the combination of multiple pieces of information from the chart; (4) Visual: Questions that rely solely on the visual aspects of the chart to extract the answer, without requiring additional reasoning or composition.

\subsection{The Impact of More Experts}

\begin{table}[h!]
    \centering
    \captionsetup{skip=0pt} 
    \caption{The impact of MoE experts numbers.}
    \label{tab:more-experts} 
        \setlength{\tabcolsep}{5pt} 
        \renewcommand{\arraystretch}{1.2} 
        \begin{tabular}{c|ccc}
        \toprule
        \#-Experts & ChartQA & Chart-to-Table & Chart-to-Text \\ 
        \midrule
        0 & 35.9 & 59.1 & 31.2 \\
        4 & 76.1 & 87.4 & 56.1 \\
        8 & 77.0 & 87.5 & 56.2 \\
        \bottomrule
        \end{tabular}
\end{table}

We experimented with using more experts, as shown in Table~\ref{tab:more-experts}. However, increasing the number of experts in the MOE architecture significantly inflates the model's parameter count, while the performance improvement is not proportional. As a result, we opted for a trade-off in the number of experts to leverage the advantages of the MOE framework fully. It is worth noting that extending the number of training steps might yield further performance gains.

\section{Limitations}

Although AskChart demonstrates competitive performance, hallucinations remain a challenge, particularly when reasoning about fine-grained visual elements within the chart. Future research could focus on enhancing the vision encoder’s capabilities, potentially through strategies such as integrating multiple encoders or employing visual token merging techniques. Moreover, the inherent limitations of large language models in managing extended context lengths pose additional constraints. Input tokens exceeding a predefined length are truncated, potentially affecting training outcomes. Investigating methods to effectively support longer context lengths could be a promising direction for improving joint representations of visual and explicit textual information.

Regarding the experimental setup, it is important to note that most of the reported results are from a single run. Pretraining is computationally intensive and costly, particularly when multiple ablation setups are considered. We believe that the results would benefit from training over a greater number of steps.

\section{Training details}

\label{app:training}

\begin{table}[tbh]
  \small
  \vspace{-10pt}
  \setlength\tabcolsep{0.75mm}
  \caption{{Training hyperparameters.}}
  \label{tab:train_detail}
  \centering
  \begin{tabular}{l|ccc}
    \toprule
            Configurations & Stage \uppercase\expandafter{\romannumeral1} & Stage \uppercase\expandafter{\romannumeral2} & Stage \uppercase\expandafter{\romannumeral3} \\
            \midrule
            Experts & -  & - & 4 \\
            Top-$k$ & -  & - & 2 \\
            \midrule
            Deepspeed & Zero2  & Zero2 & Zero2 \\
            Image resolution & \multicolumn{3}{c}{384$\times$384} \\
            Image encoder & \multicolumn{3}{c}{SigLip/384} \\
            Feature select layer & \multicolumn{3}{c}{-2} \\
            Image projector & \multicolumn{3}{c}{2 Linear layers with GeLU} \\
            Epoch & 1 & 1& 6 \\
            Learning rate &  1e-3 & 2e-5 & 2e-5 \\
            Learning rate schdule & \multicolumn{3}{c}{Cosine} \\
            Weight decay & \multicolumn{3}{c}{0.0}  \\
            Text max length & \multicolumn{3}{c}{2048} \\
            Batch size per GPU &  32 & 16 & 16 \\
            GPU & \multicolumn{3}{c}{8 $\times$ A100-80G}  \\
            Precision & \multicolumn{3}{c}{Bf16} \\
           
    \bottomrule
  \end{tabular}
\vspace{-10pt}
\end{table}

We present the training hyperparameters for all stages, as shown in Table~\ref{tab:train_detail}. We trained for 1 epoch in both of the first two stages, while in Stage III, due to the smaller dataset size, we trained for 6 epochs for appropriate total steps. The batch size was set to 256 in the first stage and 128 in the second and third stages. We utilized an image resolution of 384x384 across all three stages. Due to the excessive length of tokens extracted from the visual text, we encountered GPU out-of-memory issues in Stage III, even when using DeepSpeed's zero2\_offload mode. To address this, we employed gradient accumulation.

\section{Chart Understanding Examples}
\begin{figure*}[htbp]
	\centering
        \includegraphics[width=1\linewidth]{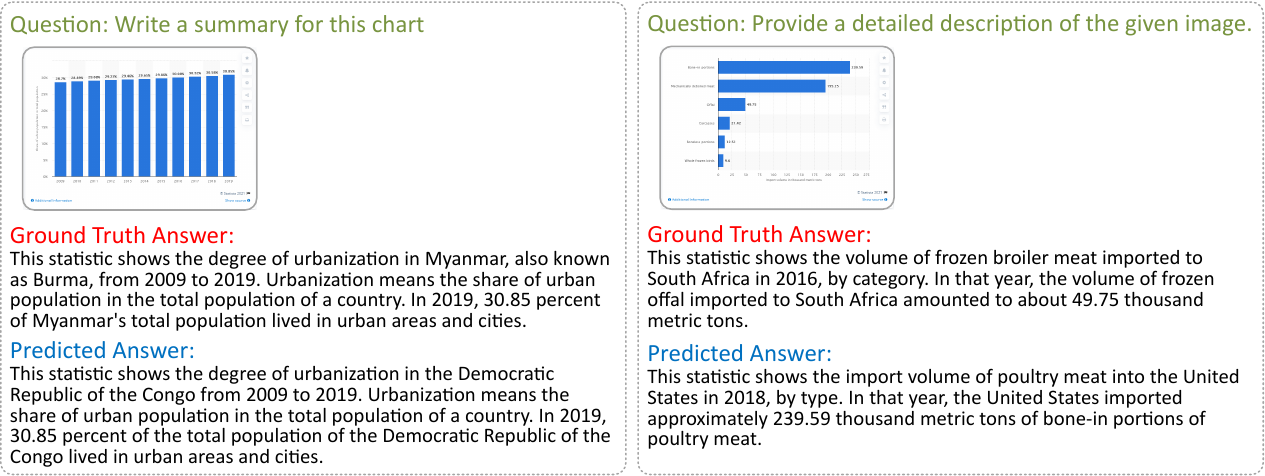}
	\vspace{-.5em}
	\caption{Examples for Chart-to-Text tasks.}
	\label{fig:chart2txt_eg}
\end{figure*}

\begin{figure*}[htbp]
	\centering
        \includegraphics[width=1\linewidth]{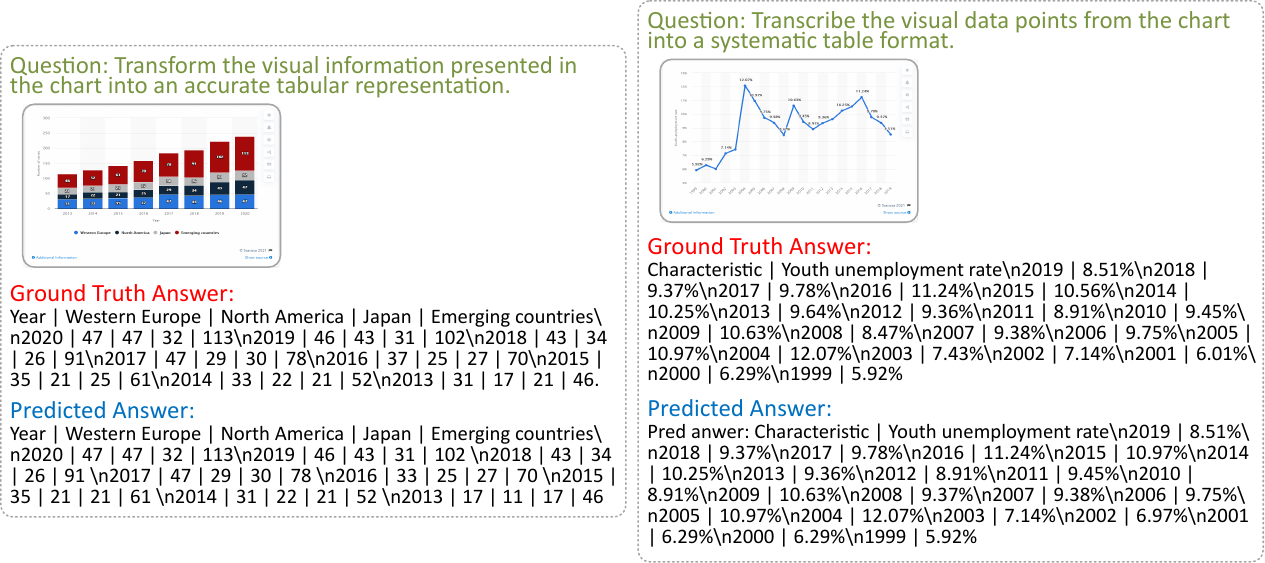}
	\vspace{-.5em}
	\caption{Examples for Chart-to-Table tasks.}
	\label{fig:chart2table_eg}
\end{figure*}

\begin{figure*}[htbp]
	\centering
        \includegraphics[width=1\linewidth]{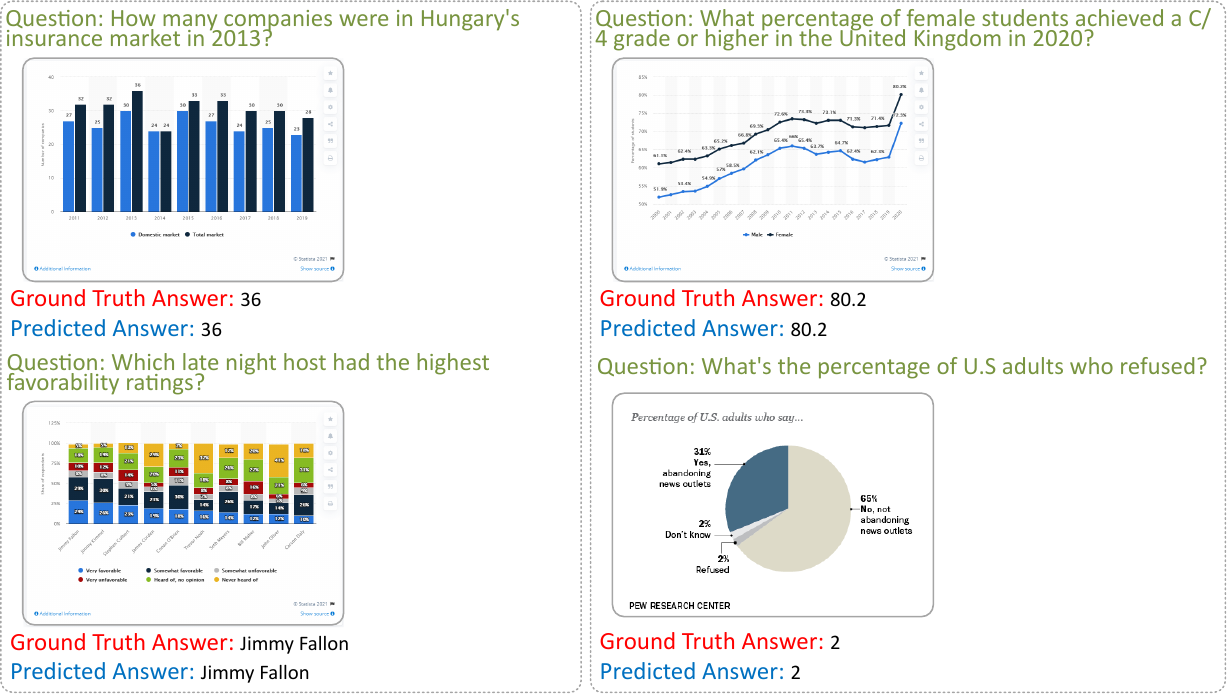}
	\vspace{-.5em}
	\caption{Examples for ChartQA tasks.}
	\label{fig:chartqa_eg}
\end{figure*}

\begin{figure*}[htbp]
	\centering
        \includegraphics[width=1\linewidth]{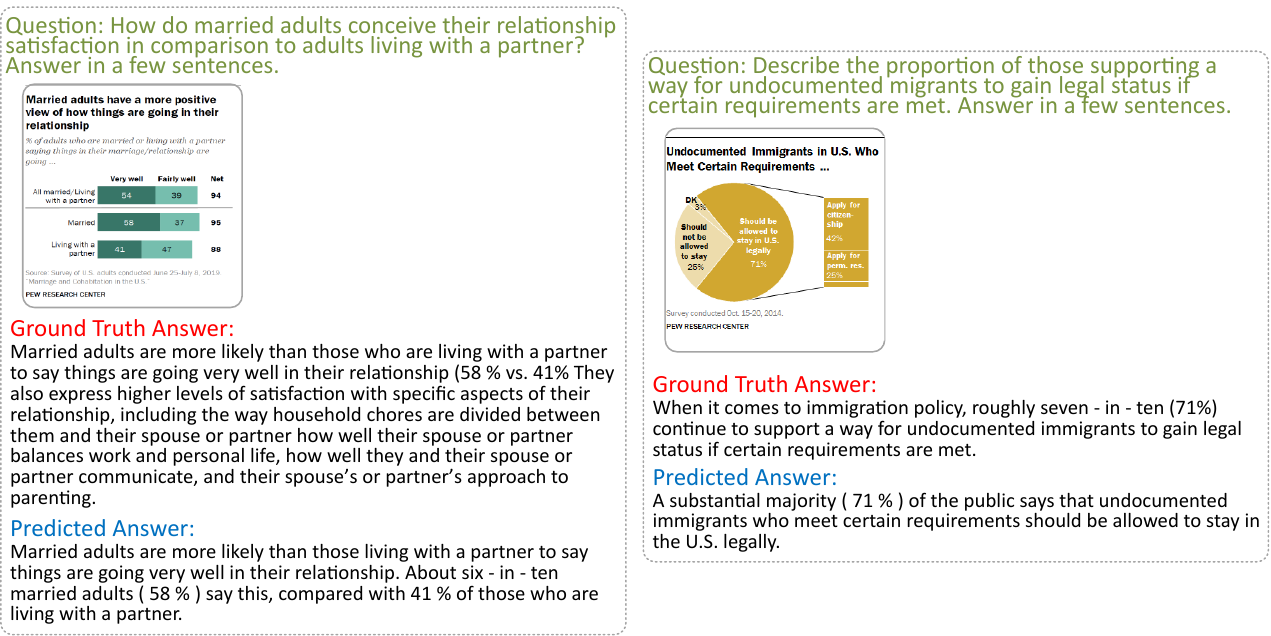}
	\vspace{-.5em}
	\caption{Examples for OpenCQA tasks.}
	\label{fig:opencqa_eg}
\end{figure*}

We below present examples for four involved chart understanding tasks: Chart-to-Text examples in Figure~\ref{fig:chart2txt_eg}, Chart-to-Table examples in Figure~\ref{fig:chart2table_eg}, ChartQA examples in Figure~\ref{fig:chartqa_eg}, and OpenCQA examples in
Figure~\ref{fig:opencqa_eg}.

\section{CoT Chart-to-Table Examples}
\label{cot_dataset}
\begin{figure*}[htbp]
	\centering
        \includegraphics[width=1\linewidth]{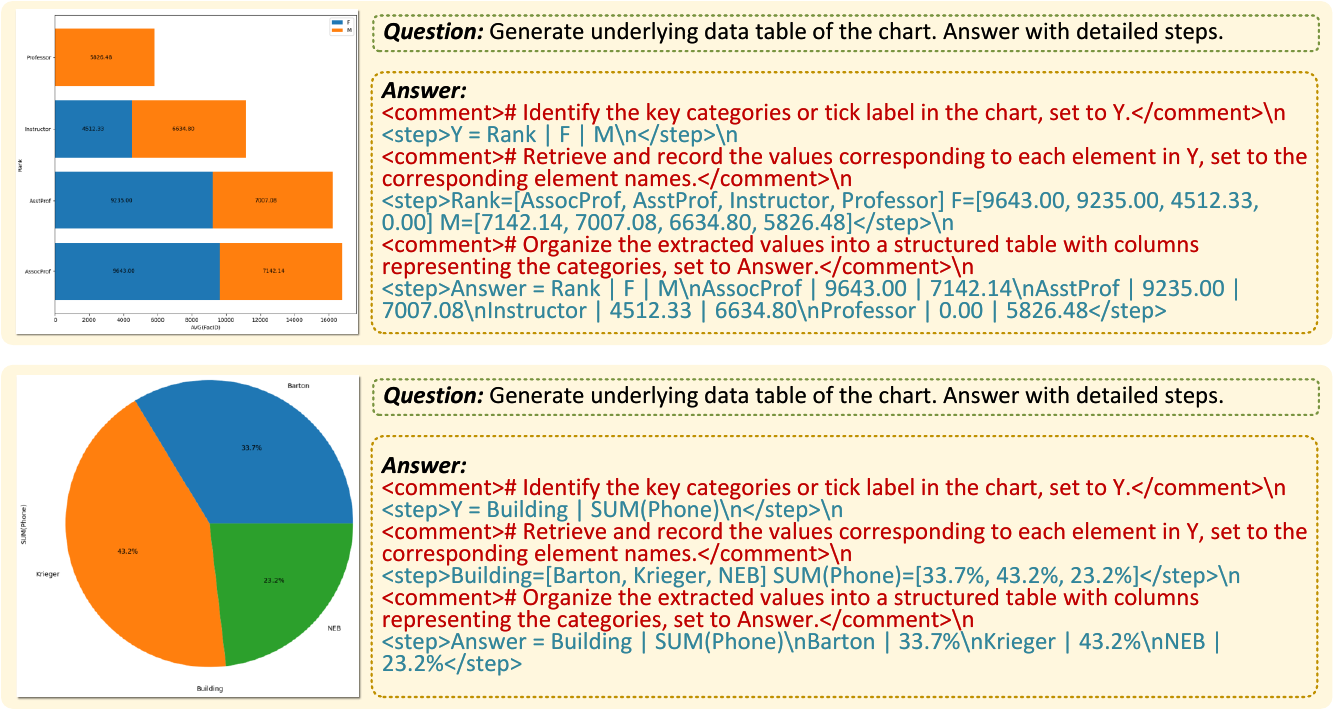}
	\vspace{-.5em}
	\caption{Two Examples of COT answers for Chart-to-Table instruction-following dataset.}
	\label{fig:cot_dataset}
\end{figure*}

Examples from the Chart-to-Table instruction-following dataset are shown in Figure~\ref{fig:cot_dataset}. The CoT (Chain-of-Thought) answer involves a multi-step reasoning process, ultimately generating the required table.

\section{Details of ChartBank}
\label{app:dataset}

In this section, we report more detailed results of \dataset.

\subsection{Instruction} \label{appsub:instruction}
We design various instruction templates to randomly select from for the chart2text and chart2table tasks, increasing expression diversity. Table~\ref{tab:inst_chart2table} and Table~\ref{tab:inst_chart2text} illustrate a portion of the instruction templates for chart2table and chart2text tasks, respectively.

\begin{table*}[htbp]
\centering
\caption{A portion of the instruction templates for the Chart-to-Table task.}
\vspace{.2em}
\renewcommand\arraystretch{1.2}
\begin{tabular}{p{13cm}}
\toprule
\textbf{Instruction Template} \\
\midrule
Extract and organize the data from the chart into a clear and concise table.\\
Create a detailed table reflecting the exact data points and categories shown in the chart.\\
Reconstruct the chart's data into a structured table, ensuring all elements are captured.\\
Translate the chart into a data table with precise values and labels as displayed.\\
Convert the charted information into a comprehensive table, including all relevant details.\\
Develop a tabular summary that encapsulates all the quantitative information from the chart.\\
Compile the data depicted in the chart into a well-organized table for easy interpretation.\\
Arrange the information contained within the chart into a methodical and detailed data table.\\
Replicate the chart's information accurately in table format, with corresponding data entries.\\
Catalog the chart data into a table, mirroring the exact figures and trends shown.\\
Transcribe the visual data points from the chart into a systematic table format.\\
\bottomrule
\end{tabular}
\label{tab:inst_chart2table}
\end{table*}

\begin{table*}[t!]
\centering
\caption{A portion of the instruction templates for the Chart-to-Text task.}
\vspace{.2em}
\renewcommand\arraystretch{1.2}
\begin{tabular}{p{13cm}}
\toprule
\textbf{Instruction Template for Brief Description} \\
\midrule
Describe the image concisely.\\
Provide a brief description of the given image.\\
Offer a succinct explanation of the picture presented.\\
Summarize the visual content of the image.\\
Give a short and clear explanation of the subsequent image.\\
Share a concise interpretation of the image provided.\\
Present a compact description of the photo's key features.\\
Relay a brief\\ clear account of the picture shown.\\
Render a clear and concise summary of the photo.\\
\midrule
\textbf{Instruction Template for Detailed Description} \\
\midrule
Describe the following image in detail.\\
Provide a detailed description of the given image.\\
Give an elaborate explanation of the image you see.\\
Share a comprehensive rundown of the presented image.\\
Offer a thorough analysis of the image.\\
Explain the various aspects of the image before you.\\
Clarify the contents of the displayed image with great detail.\\
Characterize the image using a well-detailed description.\\
Break down the elements of the image in a detailed manner.\\
Walk through the important details of the image.\\
\bottomrule
\end{tabular}
\label{tab:inst_chart2text}
\end{table*}

\subsection{Visual Prompt}

When creating a Visual Prompt dataset, we primarily follow two steps:

\paragraph{STEP1: Make questions and get bounding boxes.}Step one is to identify the relevant elements and their bounding boxes based on the question. First, we generate the corresponding queries and answers according to the predefined question templates. For example, when generating a query about finding the maximum value in a bar chart, we construct the appropriate question and locate the maximum value in the chart. Since the dataset we are using includes the bounding box coordinates for each chart element, we can identify the element corresponding to the answer by referencing the question and find the bounding box coordinates for the bar representing the maximum value.

\paragraph{STEP2: Generate Visual Prompts According to Bounding Boxes Automatically.}Step two is to automate the generation of the visual prompt using the bounding box. Here, we basically follow the rules in ViPLLaVA~\citep{vipllava}.
In our visual Prompt datasets, because we only have bounding boxes of each chart instead of pixel-level mask annotations, we only choose following visual prompt types:  arrow, triangle, ellipsis, scribble, and bounding box.
For the arrow, we make sure that the head of the arrow lies within [(-$\frac{W}{2}$, -$\frac{H}{2}$),($\frac{W}{2}$, $\frac{H}{2}$)]
 space, where $W$,$H$ are the width
 and height of the image, respectively. For the triangle, We randomly select three points within the bounding box and connect them in sequence to form a triangle. For ellipse, the lengths along the semi-major and semi-minor axes are inherited from the bounding box size, where we enlarge the
 ellipse with a ratio between [1,1.5]. For scribble, we simulate human-like drawings using Bézier curves~\cite{farin2014curves}. This process begins by randomly selecting three points within the object mask, which serve as the anchors for the quadratic Bézier curve. The generated Bézier curve is then composited onto the image using the previously mentioned alpha blending technique to produce a merged image with the scribble serving as a visual prompt. Lastly, we use bounding box coordinates to draw relevant bounding boxes as visual prompts.

\begin{figure*}[htbp]
	\centering
        \includegraphics[width=0.8\linewidth]{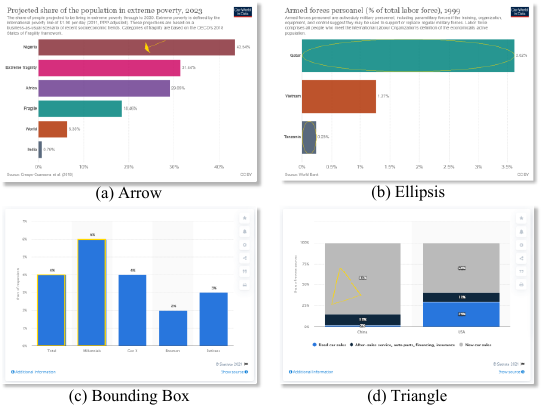}
	\caption{Four Types of Visual Prompt: Arrow, Ellipsis, Bounding Box, Triangle}
	\label{fig:visual_prompt_4eg}
\end{figure*}

Figure~\ref{fig:visual_prompt_4eg} shows examples for each type of visual prompt.

\begin{table*}[htbp]
\centering
\caption{A portion of question templates in Visual Prompt Dataset}
\vspace{.2em}
\renewcommand\arraystretch{1.2}
\begin{tabular}{p{13cm}}
\toprule
\textbf{Question Template for Reasoning} \\
\midrule
What is the sum of \{first\_x\_axis\} and \{second\_x\_axis\} in this chart?\\
What is the difference of \{first\_x\_axis\} and \{second\_x\_axis\} in this chart?
What is the mean value of \{first\_x\_axis\} and \{second\_x\_axis\} in this chart?\\
What is the total sum of all the elements in this chart?\\
What is the mean value of all the elements in this chart?\\
What is the sum of \{first\_x\_axis\} in \{first\_y\_axis\} and \{second\_x\_axis\} in \{second\_y\_axis\} in this chart?\\
What is the mean value of \{first\_x\_axis\} in \{first\_y\_axis\} and \{second\_x\_axis\} in \{second\_y\_axis\} in this chart?\\
What is the difference of \{first\_x\_axis\} in \{first\_y\_axis\} and \{second\_x\_axis\} in \{second\_y\_axis\} in this chart?\\
\midrule
\textbf{Question Template for Extremum} \\
\midrule
What is the maximum value in this bar chart?\\
What is the minimum value in this bar chart?\\
What is the maximum value in this line chart?\\
What is the minimum value in this line chart?\\
What is the maximum value in this pie chart?\\
What is the minimum value in this pie chart?\\
\midrule
\textbf{Question Template for Determine Range} \\
\midrule
What is the range of values in this bar chart?\\
What is the range of values in this line chart?\\
What is the range of values in this pie chart?\\
\midrule
\textbf{Question Template for Data Retrieval} \\
\midrule
How many bars are there in this bar chart?\\
How many pieces are there in this pie chart?\\
What is the value of \{x\_axis\} in this chart?\\
What is the value of \{x\_axis\} in \{y\_axis\}?\\
\bottomrule
\end{tabular}
\label{tab:question_temp_visual}
\end{table*}

\subsection{ChatGPT Generation prompt}

We show the question templates in the Visual Prompt Dataset in Table~\ref{tab:prompt_multi_turn}.

\begin{table*}[htbp]
\centering
\caption{Prompt ChatGPT to generate multi-turn question-answer pairs based on underlying tables of charts to construct OCR-aware Data Prompt Dataset.}
\vspace{.2em}
\renewcommand\arraystretch{1.2}
\begin{tabular}{p{13cm}}
\toprule
\textbf{\includegraphics[width=0.35cm]{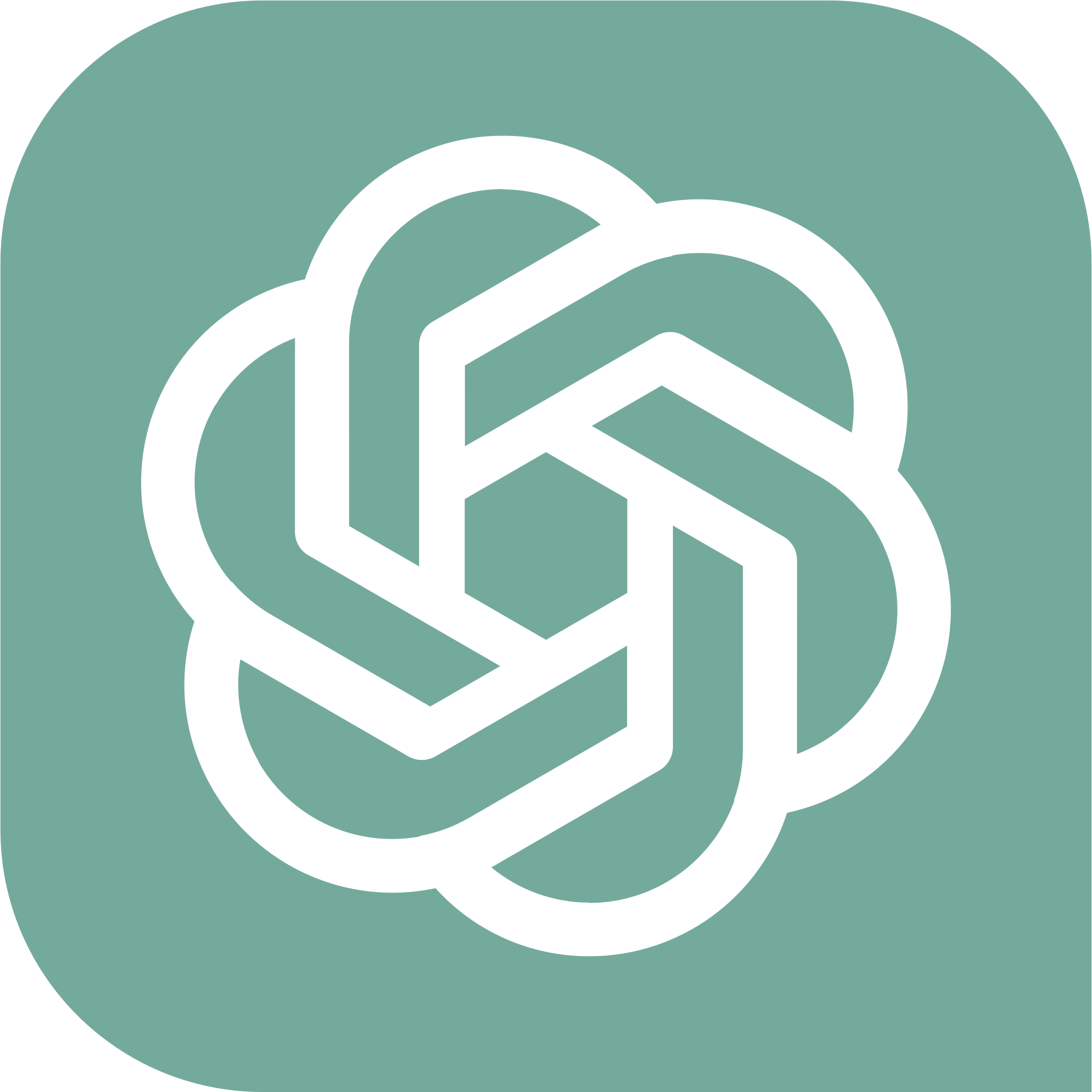} Prompt for multi-turn question-answering generation} \\
\midrule
You are an AI visual assistant that excels at chart figures. You are provided with a text description (chart summary) of a chart image and raw data values about the same chart. You don't have access to the actual image. Your task is to design question-answer pair(s) between a person (User) inquiring about the chart image and you (Assistant) responding to their questions. \\
\hline
Below are requirements for generating the question-answer pair(s):\\
- The answers should be a single word or phrase, and in a tone that a visual AI assistant is seeing the chart figure and answering the question.\\
- Ask diverse questions and give corresponding answers. Include questions asking about (1) various comparative aspects of chart image data, relationships between data points, changes over time or categories, and presence within specific ranges. (2) various numerical knowledge of chart data, including sums, differences, averages, medians, ratios, and statistical evaluations within the context of chart elements like legend labels and axis ticks or statistical measures like standard deviation, variance, and correlation and so on. \\
- The conversation should include at least 2-3 turns of questions and answers.\\
- Only include questions that have definite answers:(1) one can see in the chart figure that the question asks about and can answer confidently; (2) one can determine confidently from the chart figure that it is not in the chart figure. Do not ask any question that cannot be answered confidently. \\
- In addition, you are provided with some examples of question-answer pair(s) between a user and you(assistant).\\
\textcolor{blue}{[In context examples]}\\
The chart description: \textcolor{blue}{[Description about chart figure]} \\
The raw data: \textcolor{blue}{[Underlying data table]}\\
\bottomrule
\end{tabular}
\label{tab:prompt_multi_turn}
\end{table*}

\end{document}